\documentclass[dvipsnames,format=sigconf,anonymous=false,review=false]{acmart}

\AtBeginDocument{%
  \providecommand\BibTeX{{%
    \normalfont B\kern-0.5em{\scshape i\kern-0.25em b}\kern-0.8em\TeX}}}

\usepackage[inoutnumbered,linesnumbered,ruled,vlined]{algorithm2e} 
\usepackage{algpseudocode}
\usepackage{amsfonts}
\usepackage{amsmath}
\usepackage[english]{babel}
\usepackage{bm}
\usepackage{bbm}
\usepackage{caption}
\usepackage{changepage}
\usepackage{dsfont}
\usepackage{enumerate}
\usepackage{esvect}
\usepackage[T1]{fontenc}
\usepackage{hyperref}
\usepackage[utf8]{inputenc}
\usepackage{mathtools}
\usepackage{nccmath}
\usepackage{relsize}
\usepackage{subcaption}
\usepackage{xspace}
\usepackage{verbatim}

\newtheorem{assumption}{Assumption}

\usepackage{todonotes}

\DeclareMathOperator*{\argmax}{arg\,max}

\newcommand{\R}{{\mathbb R}}

\newcommand{\N}{{\mathbb N}}
\newcommand{\E}{\mathbb E}
\newcommand{\Prob}{\mathbb{P}}
\newcommand\eps{\varepsilon}
\newcommand{\CA}{{\mathcal A}}

\newcommand{\CC}{{\mathcal C}}
\newcommand{\CD}{{\mathcal D}}
\newcommand{\CE}{{\mathcal E}}

\newcommand{\Bin}{{\mathrm{Bin}}}
\newcommand{\sss}[1]{{\scriptscriptstyle #1}}
\newcommand{\plus}{\mathrm{plus}}

\newcommand{\ind}[1]{{\mathbbm 1\{#1\}}}

\newcommand{\Tcom}{{T^\sss{\mathrm{com}}}}

\newcommand{\Tcomall}{{T^\sss{\mathrm{com}}_{k^\ast,\lambda,d,p}}}
\newcommand{\Tcomdyn}{{T^\sss{\mathrm{com,dy}}}}
\newcommand{\Tcomdynall}{{T^\sss{\mathrm{com,dy}}_{k^\ast,\lambda,d,p}}}
\newcommand{\Tplus}{{T^\sss{\mathrm{plus}}}}

\newcommand{\Tplusall}{{T^\sss{\mathrm{plus}}_{k^\ast,\lambda,d,p}}}

\newcommand{\Tdis}{{T^\sss{\disOM}}}
\newcommand{\TdisC}{{T_\CC^\sss{\disOM}}}
\newcommand{\TdisD}{{T_\CD^\sss{\disOM}}}
\newcommand{\TdisDk}{{T_{\CD,k}^\sss{\disOM}}}
\makeatletter
\newcommand*{\textoverline}[1]{$\overline{\hbox{#1}}\m@th$}
\makeatother

\usepackage[utf8]{inputenc}

\newcommand{\loge}{\log_{\eta}}
\newcommand{\oea}{${(1+1)}$~EA\xspace}
\newcommand{\ocl}{${(1,\texorpdfstring{\lambda}{λ})}$~EA\xspace}
\newcommand{\mcl}{${(\mu , \lambda)}$~EA\xspace}
\newcommand{\opl}{${(1+\texorpdfstring{\lambda}{λ})}$~EA\xspace}

\newcommand{\onemax}{\textsc{OneMax}\xspace}
\newcommand{\zeromax}{\textsc{ZeroMax}\xspace}
\newcommand{\OM}{\textsc{OM}\xspace}
\newcommand{\ZM}{\textsc{ZM}\xspace}
\newcommand{\disOM}{\textsc{disOM}\xspace}
\newcommand{\dydisOM}{\textsc{dyDisOM}\xspace}
\newcommand{\jump}{\textsc{Jump}\xspace}
\newcommand{\cliff}{\textsc{Cliff}\xspace}

\newcommand{\pimp}{p_{\text{imp}}}
\newcommand{\pimpone}{p_{\text{imp},1}}

\newif\iflong
 \longtrue
\newif\ifshort
 \shortfalse 
\newcommand{\inLongVersion}[1]{\iflong #1\fi}
\newcommand{\inShortVersion}[1]{\ifshort #1\fi}

\inLongVersion{
\settopmatter{printacmref=false} 
\renewcommand\footnotetextcopyrightpermission[1]{} 
\pagestyle{plain} 
\usepackage{balance} 
}

\makeatletter
\def\blfootnote{\xdef\@thefnmark{}\@footnotetext}
\makeatother

\inShortVersion{
\copyrightyear{2023}
\acmYear{2023}
\setcopyright{acmlicensed}\acmConference[GECCO '23]{Genetic and Evolutionary Computation Conference}{July 15--19, 2023}{Lisbon, Portugal}
\acmBooktitle{Genetic and Evolutionary Computation Conference (GECCO '23), July 15--19, 2023, Lisbon, Portugal}
\acmPrice{15.00}
\acmDOI{10.1145/3583131.3590488}
\acmISBN{979-8-4007-0119-1/23/07}
}
\inLongVersion{
\setcopyright{acmlicensed}\acmConference[An extended abstract appears at GECCO]{Genetic and Evolutionary Computation Conference}{2023}{}
}

\begin{document}

\title{Comma Selection Outperforms Plus Selection on OneMax with Randomly Planted Optima\inShortVersion{$^\dagger$}}


\author{Joost Jorritsma}
\affiliation{%
  \institution{Eindhoven University of Technology}
  \city{Eindhoven}
  \country{Netherlands}
}

\author{Johannes Lengler}
\affiliation{%
  \institution{ETH Z\"urich}
  \city{Z\"urich}
  \country{Switzerland}
}

\author{Dirk Sudholt}
\affiliation{%
  \institution{University of Passau}
  \city{Passau}
  \country{Germany}
}

\renewcommand{\shortauthors}{Jorritsma, Lengler, Sudholt}

\begin{abstract}
It is an ongoing debate whether and how comma selection in evolutionary algorithms helps to escape local optima. We propose a new benchmark function to investigate the benefits of comma selection: \onemax with randomly planted local optima, generated by frozen noise. We show that comma selection (the \ocl) is faster than plus selection (the \opl) on this benchmark, in a fixed-target scenario, and for offspring population sizes~$\lambda$ for which both algorithms behave differently. For certain parameters, the \ocl finds the target in $\Theta(n \ln n)$ evaluations, with high probability (w.h.p.), while the \opl w.h.p.\ requires almost $\Theta((n\ln n)^2)$ evaluations.

We further show that the advantage of comma selection is not arbitrarily large: 
 w.h.p.\ comma selection outperforms plus selection at most by a factor of $O(n \ln n)$ for most reasonable parameter choices. 
We  develop novel methods for analysing frozen noise and give powerful and general fixed-target results with tail bounds that  are of independent interest.
\end{abstract}
\maketitle



\section{Introduction}

\inShortVersion{\blfootnote{$^\dagger$Extended abstract. Full proofs can be found at~\cite{???}. 
}}
Evolutionary Algorithms (EAs) are optimisation heuristics that are very flexible. An important aspect of EAs are their selection strategies. \emph{Elitist} strategies like plus selection always maintain the best-so-far search point in the population.\footnote{Some authors define elitism in a stronger way, so that the whole population must consist of the best-so-far search points~\cite{doerr2017introducing}. } While elitism is helpful to exploit the best-so-far solution, a common concern is that the algorithm might get stuck in local optima. The \emph{escape hypothesis} posits that non-elitism might help in such cases~\cite{dang2021escaping}. Indeed, the disadvantage of elitism can be measured by the \emph{elitist black-box complexity} of a problem, and some (artificial) problems show an exponential penalty of elitism~\cite{doerr2017introducing,dang2021non,dang2021escaping}. 

A popular non-elitist selection strategy is \emph{comma selection}, in which the parent(s) are not allowed to compete for survival. Despite its popularity, it is still unclear how good comma selection is at helping with escaping local optima. Some deceptive landscapes, in which other non-elitist mechanisms can help, can still deceive comma selection~\cite{dang2021non}. A working group at a 2022 Dagstuhl seminar ``came to the conclusion that there is a gap between theory and practice as we are lacking convincing examples (apart from \cliff) where comma selection provably helps, whereas in
practice comma strategies seem to be quite popular to escape from local optima''~\cite{dagstuhl22081theory}. 

The main theoretical results for comma strategies at local optima are for the benchmarks \jump and \cliff. It is known that comma selection in the \mcl does not work better than plus selection on the \jump function, for arbitrary population sizes $\mu$ and $\lambda$~\cite{doerr2020does}. For the \cliff function, there is a choice of $\lambda$ such that the \ocl ``only'' takes time $O(n^{3.97\ldots})$, while the plus strategy needs exponential time for any $\lambda$~\cite{jagerskupper2007plus,rowe2014choice,hevia2021a}. However, the optimisation time of the \ocl is still rather high (albeit polynomial), and the dependence on $\lambda$ is rather tricky. There are some promising efforts to develop self-adjusting mechanisms that can adapt $\lambda$ during the run of the algorithm, but these come with their own pitfalls~\cite{hevia2021a,kaufmann2022self,kaufmann2023onemax,Hevia2022}. In particular, the optimisation time for \cliff can be reduced to $O(n\ln n)$ by a self-adjusting mechanism which resets $\lambda$ periodically, but this mechanism needs to be well-aligned with the problem~\cite{hevia2021a}.

Arguably, \cliff (and also \jump) captures a rather specific situation that might be atypical for local optima. We will not give the definition of \cliff, but only describe the atypical situation. When the algorithm is in a local optimum $x$, and accepts a Hamming neighbour $y$ that is closer to the optimum (``down the cliff''), then \emph{most neighbours of $y$ have the same fitness as $x$} and thus are ``back up on the cliff''. In particular, a random walk without selective pressure would likely lead back to the local optimum (or an equivalent search point). It is rather hard to imagine this situation in practice, where at or near an optimum (global or local), random walks typically increase the distance from the optimum and decrease fitness. This is not just a coincidental feature of the \cliff function; the analysis of this phenomenon is at the heart of all runtime analyses of \cliff. Thus, \cliff (and \jump) might not be the best test function for understanding local optima. 

One issue with the \ocl is that one needs to find a compromise between two trends: if $\lambda$ is too large, then every generation contains a clone of the parent, so the \ocl just imitates an elitist algorithm. But if $\lambda$ is too small then the algorithm can not cope with situations in which improvements are hard to find. If the goal is to find a global optimum, then it is hard to balance those two aspects. However, it is less problematic in fixed-target optimisation, which is why we study this setting. We will discuss this point in more detail in Section~\ref{sec:setup}.

\subsection{Distorted OneMax}
We introduce an alternative model of local optima, which we call \textsc{distorted OneMax} or $\disOM = \disOM_{d,p}$ for short. It could also be called ``\onemax with planted local optima'' or ``\onemax with frozen Bernoulli noise''. We start with the \onemax function and two real-valued parameters $p\in [0,1]$ and $d > 0$. Then for each search point $x$, with probability $p$ we increase its fitness by $d$, independently of the other search points. Hence, we artificially ``plant'' a local optimum in $x$. (It does not always need to be a local optimum since a fitter neighbour of $x$ could also be distorted.) Note that the distortion is part of the fitness function, which is static. Hence, if an algorithm evaluates the same search point $x$ several times, it will always detect the same fitness $f(x)$. This is different from models with noisy fitness evaluations, in which several queries for the same search point can give different fitness values. In independent, concurrent work~\cite{FriedrichKNR22} another frozen noise model was recently studied for the compact Genetic Algorithm~cGA on \onemax. There, Gaussian noise was added to all search points.

\subsection{Main Results}\label{sec:results}

We study the fixed-target performance with fitness target $n-k^{\ast}$ of the \ocl and the \opl on distorted \onemax with parameters $p\in[0,1]$ and $d>1$. We will explain those choices in more detail in Section~\ref{sec:setup} below. We denote by $\Tcom = \Tcomall$ and $\Tplus = \Tplusall$ the number of function evaluations until the \ocl and the \opl find a search point of fitness at least $n-k^{\ast}$ on $\disOM_{d,p}$, respectively.

We give matching upper and lower bounds on $\Tcom$ and $\Tplus$ in Theorem~\ref{thm:main} below for a wide range of parameters $k^\ast$, $\lambda$, $d$, $p$, which hold with high probability.\footnote{\emph{With high probability} (w.h.p.) means with probability $1-o(1)$ as $n\to \infty$.} For those parameters $\Tplus$ is by a factor $1/p$ larger than  $\Tcom$. In particular, we show that there are parameters for which comma selection reduces the runtime from nearly quadratic to quasi-linear.
Since the assumptions on the parameters are a bit technical, we state them together with a discussion in Assumption~\ref{ass:main} in Section~\ref{sec:setup} below. Intuitively, the parameters $\lambda$ and $k^\ast$ must be chosen such that the \ocl efficiently reaches fitness target $n-k^\ast$, but it does not create a clone of the parent in each generation. We state our main theorem. 
\begin{theorem}\label{thm:main}
    Under Assumption~\ref{ass:main} on the parameters below, with high probability
    \begin{align}\label{eq:maincom}
        \Tcomall &= \Theta\big( n\ln n\big),\\
    \label{eq:mainplus}
        \Tplusall & =\Theta(n\ln(n)/p).
    \end{align}
    For any $p$ such that $p=\omega(1/(n\ln n))$ and $p= n^{-\Omega(1)}$ there are $k^\ast$, $\lambda$ and $d$ such that Assumption~\ref{ass:main} is satisfied. Thus,~\eqref{eq:maincom} and~\eqref{eq:mainplus} may differ by a factor of almost $n \ln n$.
    
\end{theorem}

It may seem like an unimportant quirk that we have used a w.h.p.\ statement instead of expectations, but this is not so. Indeed, the \textsc{distorted OneMax} function easily leads to regimes in which expectations are completely meaningless, since they are dominated by events of tiny probability which contribute gigantic terms to the expectation. This is a known phenomenon, see~\cite{doerr2017introducing} for an in-depth discussion in the context of elitist black-box complexity. In our situation, we give the following proposition as example. We remark that those parameters are outside of the regimes of Assumption~\ref{ass:main}.
\begin{proposition}\label{prop:Markov-Vegas}
 For $p = 2^{-n}$, $d=n-0.5$, $k^{\ast} = 0$ and $\lambda = 3\ln n$, 
 \begin{align}\label{eq:LasVegas}
 \E\big[\Tcomall\big]  = O(n \ln n) \quad\text{and}\quad \E\big[\Tplusall\big] = n^{\Omega(n)},    
 \end{align}
but with high probability
 \begin{align}\label{eq:MonteCarlo}
 \Tcomall  = O(n \ln n) \quad\text{and}\quad \Tplusall = O(n\ln n).   \end{align}
\end{proposition}
\inShortVersion{The proof of Proposition~\ref{prop:Markov-Vegas} is omitted.}
Following the terminology of~\cite{doerr2017introducing} for black-box complexity, we also call the expected times in~\eqref{eq:LasVegas} the \emph{Las Vegas runtimes}, and for $r\in [0,1]$ we call the $r$-\emph{Monte Carlo runtime} the time until the algorithm finds the target with probability at least $1-r$. Proposition~\ref{prop:Markov-Vegas} states that the Las Vegas runtime and the $r$-Monte Carlo runtime (for any constant $r$) differ dramatically for the \opl. In general, Monte Carlo runtimes are more informative since they make a statement about \emph{typical} outcomes.\footnote{Usually, there is another reason to prefer Monte Carlo runtimes, since with good Monte-Carlo runtimes we can restart the algorithm if a run gets stuck. However, the situation here is a bit more subtle, since \disOM is a randomized function. Thus, there are two forms of randomness: one from the random choice of the fitness function, and one from the random decision of the algorithms. If the long expected runtime comes from an atypical fitness function, the problem is not solved by restarting the algorithm. However, it is  easy to find a \emph{fixed} function for which Proposition~\ref{prop:Markov-Vegas} still holds, for example the \onemax function with a single planted local optimum of value $n-1$ at the all-zero string. This is implicitly shown in the proof of Proposition~\ref{prop:Markov-Vegas}.}

Theorem~\ref{thm:main} gives a factor $1/p$ between the comma and the plus strategy that is arbitrarily close to $n \ln n$. The next theorem shows that this is the largest possible factor for Monte Carlo runtimes if the \ocl is efficient on \disOM. Note that the factor for Las Vegas runtimes can be huge by Proposition~\ref{prop:Markov-Vegas}. 
\begin{theorem}\label{thm:largest-factor}
    Let $C,\eps>0$ be constants. Assume that the parameters $\lambda \ge 1$, $k^\ast\in[n^{\varepsilon}, n/6]$, $p\in [0,1]$ and $T \in[1,n^C]$ (all possibly depending on $n$) are such that with high probability
    \begin{align}\label{eq:largest-factor-com}
        \Tcomall \le T.
    \end{align}
     Then there exists a constant $C'>0$ such that with high probability
    \begin{align*}
        \Tplusall\!\! \le\! 
        \begin{dcases}
            \!T,&\!\!\text{if }p\cdot T=o(1) \text{ or }\lambda>C'\ln(n),\\
            \!O\big(n(\lambda\! +\! \ln(n/k^\ast))\big),\!\!&\!\!\text{if }p\cdot n(\lambda + \ln(n/k^\ast)) = o(1), \\
            \!O\big(T/p \big),&\!\!\text{otherwise, provided }\lambda\!=\!O(\ln(n/k^\ast)).
        \end{dcases}
    \end{align*}
    Moreover, in all three cases 
    $\Tplus=O(T\cdot n\ln n)$ w.h.p.
\end{theorem}
We believe that the condition on $\lambda$ in the third case is not needed, and hope to remove this condition in future work.
We emphasize that the conditions in Theorem~\ref{thm:largest-factor} are much more general than Assumption~\ref{ass:main} and also cover many degenerate parameter settings. For example, for small $\lambda$ the \ocl may be inefficient and not have runtime $\Theta(n\ln n)$. 
We caution that Theorem~\ref{thm:largest-factor} is far from trivial. We do make heavy use of the condition $k^\ast \ge n^\eps$ in the proof. Moreover, we believe that the statement would be wrong if the \opl would break fitness ties in favour of the parent. We discuss those matters in more detail in Section~\ref{sec:plus-upper}.

\subsection{Parameter Setup}\label{sec:setup}
We will now explain which regimes are reasonable to consider for the parameters $k^\ast,\lambda,d,p$. Note that all of $d=d(n)$, $k^\ast = k^\ast(n)$, $p=p(n)$ and $\lambda = \lambda(n)$ may depend on $n$. 

We will use the abbreviation $\eta:= e/(e-1)$. This is helpful since the probability that a mutation is not identical to the parent (is not a \emph{clone}) is $\approx 1-1/e = \eta^{-1}$. We write $q \coloneqq \eta^{-\lambda}$ for the (approximate) probability to have no clone in the offspring population. This is a central quantity, since it is the probability of escaping from a local optimum in the \ocl. It has been known for decades that if $\lambda \ge C\ln n$ for a large $C$, then the \ocl mimics an elitist algorithm because then $q = n^{-C \cdot \ln(\eta)} \approx n^{-0.66 C}$~\cite{jagerskupper2007plus,lehre2011fitness}. For example, if $C\ge 4$ then $q =o(n^{-2})$ and w.h.p.\ the parent will be cloned in all of the first $n^{2}$ generations. Hence, the \ocl just behaves as the \opl in this regime. Since we are interested in potential differences between those two algorithms, we will thus consider regimes where $q \ge n^{-1+\eps}$, or equivalently $\lambda \le (1-\eps)\loge n $, since this is the regime where the two algorithms behave differently~\cite{lehre2011fitness}. Note that this choice is not wise when the algorithm is supposed to find the optimum, since the \ocl with $\lambda \le (1-\eps)\loge n$ is inefficient in finding the optimum of any function with unique optimum~\cite{rowe2014choice}. However, such a $\lambda$ is fine for fixed-target optimisation, i.e., if we are interested in finding a search point of fitness at least $n-k^\ast$, as long as $\lambda \ge (1+\eps)\loge(n/k^\ast)$~\cite{antipov2019efficiency}, or equivalently $q \le (k^\ast/n)^{1+\eps}$. Thus, we require $(1+\eps)\loge(n/k^\ast) \le \lambda \le (1-\eps)\loge n$ for some constant $\eps >0$. This range is non-empty if $k^\ast = n^{\Omega(1)}$, so we will make this restriction. This also avoids some complications at the optimum, since the optimum of \disOM may be not at $\vec 1 = (1,\ldots, 1)$, but at a distorted point whose fitness exceeds $n$. For this reason we will assume $d \le k^\ast$, so that the target fitness $n-k^\ast$ cannot be achieved by search points at distance larger than $2k^\ast$ from $\vec 1$.

Furthermore, the \opl with $\lambda = O(\ln n)$ needs $O(n \ln n)$ fitness evaluations to optimize \onemax. If $p= o(1/(n\ln n))$, then w.h.p.\ the \opl will not encounter any distorted search points before finding the optimum. Thus, we may ignore the case $p= o(1/(n\ln n))$. We will restrict ourselves a little bit more, and assume $p = \omega(1/n)$. 
Finally, we will make two more assumptions for technical simplicity.
Firstly, we require $p = o(k^\ast/n)$. This ensures that close to the target fitness $n-k^\ast$, the probability $\Theta(k^\ast/n)$ that an offspring is closer to $(1,\ldots, 1)$ dominates the probability $p$ that the offspring is distorted. 
\inShortVersion{The regime $p = \omega(k^\ast/n)$ is rather different, and we leave the study of this regime for future work. }
\inLongVersion{The regime $p = \omega(k^\ast/n)$ is rather different because even if the comma strategy escapes from a local maximum into a non-distorted point, it will likely return to a distorted point before having the chance to make an improvement. We leave the study of this regime for future work. }
Secondly, we assume that $q = \omega(p\lambda)$. Note that $p\lambda$ is roughly the probability of sampling a distorted offspring in one generation, while $q$ is the probability of escaping a local optimum in the \ocl. Thus, we assume that escaping is to be more likely than sampling another local optimum. This condition simplifies the analysis in some places, but we don't believe it is actually needed, and we hope that we can remove it in future work. In fact, we will require the slightly stronger condition $q \ge p^{1-\eps}$, or equivalently $\lambda \le (1-\eps)\loge(1/p)$. This is stronger than $q = \omega(p\lambda)$ since the other conditions already imply $p=n^{-\Omega(1)}$ and $\lambda = \Theta(\ln n)$. 
Summarizing, we will make the following assumption for our main theorem below.
\begin{assumption}\label{ass:main}
Let $q:= \eta^{-\lambda}$ for $\eta := e/(e-1)$, and let $\eps >0$ be any constant. We assume $k^\ast = n^{\Omega(1)}$ and $k^\ast = n^{1-\Omega(1)}$, $p =\omega(1/(n\ln n))$, and
\begin{align}\label{eq:range-of-q}
p^{1-\eps} \le q \le (k^\ast/n)^{1+\eps}.
\end{align}
Finally, we assume that $d\in[(1+\eps)\ln(n/p)/\ln(n/k^\ast), k^\ast]$.
\end{assumption}
We have already motivated the assumptions on $p$, $k^{\ast}$ and $\lambda$, and the upper bound on $d$. The lower bound on $d$ will come out of the proof of the lower bound on $\Tplus$. By the assumptions on $k^\ast$ and $p$ we have $\ln(n/p) = \Theta(\ln n)$ and $\ln(n/k^\ast) = \Theta(\ln n)$, so the lower bound on $d$ is just a constant. 

Note that we can write~\eqref{eq:range-of-q} equivalently as a condition on $\lambda$:
\begin{align}\label{eq:range-of-lambda}
(1+\eps) \loge(n/k^\ast) \le \lambda \le (1-\eps)\loge(1/p).
\end{align}
Also, Assumption 1 implies $q=n^{-\Theta(1)}$, and thus for some $\delta >0$,
\begin{equation}
p\le q^{1+\varepsilon/(1-\varepsilon)}\le qn^{-\delta}\le n^{-2\delta},\quad\mbox{and} \quad p\le q \le k^\ast/n^{1+\delta}.\label{eq:ass-rewritten}
\end{equation}
We discuss briefly the possible ranges of the parameters. The values of $p$, $q$, and $k^\ast/n$ are coupled by~\eqref{eq:range-of-q}, but $p$ can be arbitrarily close to $1/(n \ln n)$ and $k^\ast$ can take a value $n^c$ for a constant $c<1$ that is arbitrarily close to $1$. Hence, for any constant $0<c<1$ we may set any one of the three values $p$, $q$, or $k^\ast/n$ to $n^{-c}$ and still satisfy Assumption~\ref{ass:main} by choosing the other two values appropriately. As discussed above, values of $p$ or $q$ much smaller than $n^{-1}$ do not lead to interesting regimes, because respectively the algorithm does not encounter distorted points or it mimics a plus strategy. The restrictions on $q$ always determine $\lambda$ up to constant factors, where the interval may be more or less narrow depending on $p$ and $k^\ast$. This discussion also implies the second statement of Theorem~\ref{thm:main}, that we may choose parameters yielding a factor of more than $n$ between $\Tcom$ and $\Tplus$.
\begin{proof}[Parameters yielding quasi-linear factor in Theorem~\ref{thm:main}]\mbox{\hspace*{0.5pt}} 
Let $\delta >0$ and $p=p(n)$ such that $p\le n^{-\delta}$ and $p = \omega(1/(n\ln n))$. Then Assumption~\ref{ass:main} is satisfied by setting $q := n^{-\delta/2}$, $k^\ast := n^{1-\delta/4}$, and $d$ as a sufficiently large constant. 
\end{proof}

\section{Notation and Preliminaries}

\paragraph{General Notation.} We write $[n] := \{1,\ldots,n\}$. We denote search points by $x=(x_1,\ldots,x_n)\in\{0,1\}^n$, and the \onemax value of $x$ is $\OM(x) := \sum_{i\in [n]}x_i$. We denote by $\vec 1 = (1,\ldots,1)$ and $\vec 0 = (0,\ldots,0)$ the unique search points with $\OM(\vec 1) =n$ and $\OM(\vec 0) =0$. For $x,y\in \{0,1\}^n$, the \emph{Hamming distance} $H(x,y)$ of $x$ and $y$ is the number of positions $i\in[n]$ such that $x_i \neq y_i$. We call $y$ a (Hamming) \emph{neighbour} of $x$ if $H(x,y)=1$. We set $\eta:= e/(e-1)$, and we denote by $\log_2 n$ the binary logarithm of~$n$, by $\ln n$ the natural logarithm of $n$, and by $\log_\eta n := \ln n/\ln\eta$ the logarithm with base $\eta$. 

For an event $\CE$ we denote by $\ind\CE$ the indicator variable of $\CE$, i.e., $\ind\CE = 1$ if $\CE$ occurs and $\ind\CE = 0$ otherwise. 

\paragraph{\textsc{Distorted OneMax.}} We start with a formal definition of the \textsc{distorted OneMax} function $\disOM: \{0,1\}^n\to \R_{\ge 0}$. We partition the search space $\{0,1\}^n$ into two sets $\CC$ and $\CD$ of ``clean'' and ``distorted'' points, respectively, where for each $x\in \{0,1\}^n$ we have 
\begin{equation}
    x\in \CD\text{ with probability }p, \qquad x\in\CC\text{ otherwise},
\end{equation}
independently of the other points. We define the \textsc{distorted OneMax} function $\disOM = \disOM_{d,p}$ as  
\begin{equation}
    \disOM(x) := \OM(x) + d\cdot \ind{x\in\CD}.
\end{equation}

\begin{algorithm}[thb]
\caption{$(1,\lambda)$~EA for maximizing $f$ to target $n-k^\ast$.}\label{alg:comma}
\SetKwInput{Init}{Initialization}
\SetKwInput{Mut}{Mutation}
\SetKwInput{Sel}{Selection}
\SetKwInput{Opt}{Optimization}
\SetKwInput{Upd}{Update}
\Init{$t=0$; pick $x^\sss{(0)}$ uniformly at random from $\{0,1\}^n$.}
\Opt{
	\While{$f(x^\sss{(t)}) \le n-k^\ast$}
		{
		\Mut{
			\For{$j \in \{1,\dots,\lambda \}$}
				{
				$y^\sss{(t,j)}\leftarrow$ mutate$(x^\sss{(t)})$ by flipping each bit of $x^\sss{(t)}$ independently with prob. $1/n$\;
				}}
		\Sel{
			 Let $y^\sss{(t)} = \argmax\{f(y^\sss{(t,1)}), \dots, f(y^\sss{(t,\lambda)})\}$, breaking ties uniformly at random\;}
		\Upd{$x^\sss{(t+1)}=y^\sss{(t)}$;\qquad
			$t\leftarrow t+1$;}
		}
}
\end{algorithm}
\begin{algorithm}[tbh]
\caption{$(1+\lambda)$~EA for maximizing $f$ to target $n-k^\ast$.}\label{alg:plus}
\SetKwInput{Init}{Initialization}
\SetKwInput{Mut}{Mutation}
\SetKwInput{Sel}{Selection}
\SetKwInput{Opt}{Optimization}
\SetKwInput{Upd}{Update}
\Init{$t=0$; pick $x^\sss{(0)}$ uniformly at random from $\{0,1\}^n$.}
\Opt{
	\While{$f(x^\sss{(t)}) \le n-k^\ast$}
		{
		\Mut{
			\For{$j \in \{1,\dots,\lambda \}$}
				{
				$y^\sss{(t,j)}\leftarrow$ mutate$(x^\sss{(t)})$ by flipping each bit of $x^\sss{(t)}$ independently with prob. $1/n$\;
				}}
		\Sel{
			 Let $y^\sss{(t)} = \argmax\{f(y^\sss{(t,1)}), \dots, f(y^\sss{(t,\lambda})\}$, breaking ties uniformly at random\;}
		\Upd{
			 \algorithmicif\ {$f(y^\sss{(t)}) \ge f(x^\sss{(t)})$}\ \algorithmicthen\ $x^\sss{(t+1)}=y^\sss{(t)}$; 
			 \algorithmicelse\ $x^\sss{(t+1)}=x^\sss{(t)}$;\qquad
			 $t \leftarrow t+1$;
			}
		}
}
\end{algorithm}

\paragraph{Algorithms.} In Algorithms~\ref{alg:comma} and~\ref{alg:plus} we give the pseudocode for fixed-target optimisation of a fitness function $f$ with the \ocl and the \opl respectively. The \emph{running time} is the number of function evaluations until the target is met. Note that in our context of $\disOM$, the target will always be to reach fitness at least $n-k^\ast$. Throughout the paper, we assume that the mutation rate is $1/n$.

We recall that $\Tcom = \Tcomall$ and $\Tplus = \Tplusall$ are the number of function evaluations until the \ocl and the \opl respectively find a search point of fitness at least $n-k^{\ast}$ on $\disOM_{d,p}$. 
By the \emph{time} that an algorithm spends, we refer to the number of function evaluations. So when the algorithm runs for time $T$, then it runs for $\lceil T/\lambda \rceil$ generations.

\section{General Tools}

In this section we collect some lemmas which we need for our results, but which may be useful in other contexts as well. Section~\ref{sec:prop-com} collects basics about the \ocl\inLongVersion{, in particular bounds on the drift and on the probability of producing a clone}. In Section~\ref{sec:domination}, we prove that the \opl is the most efficient algorithm on \onemax (in the sense of stochastic domination) among all $(1+\lambda)$ algorithms with the same fixed mutation rate. This also holds in fixed-target settings. As a corollary, we obtain that the \opl on \onemax is the fastest algorithm to reach a fixed Hamming distance from the optimum among all $(1+\lambda)$ algorithms and all fitness function with a unique global optimum. 
In Section~\ref{sec:fixed-target} we show high-probability upper and lower runtime bounds for the \opl and \ocl on \onemax with prescribed start and target fitness. 
The results in this section will not come as a big surprise for experts since similar, but more specific, statements were known before. However, this is the first time they are proven in such generality.

\subsection{Properties of the \ocl}\label{sec:prop-com}
\inShortVersion{The following two lemmas are standard. They confirm that the probability of not creating a clone is $\approx q$, and that it is unlikely to have any mutation with more than $c\ln n$ bit-flips within the first $n^2$ generations. We remark that the following lemma is numbered~3.2 to maintain consistency with the full version~\cite{???}.}
\inLongVersion{
The following lemma summarizes known results on transition probabilities and expectations from the literature~\cite{rowe2014choice,hevia2021self,Bossek2021a}.
\begin{lemma}
\label{lem:transition-probabilities}
Let $X_t = \OM(x^{(t)})$ be the current fitness of the \ocl on \onemax and let $\Delta_k \coloneqq (X_t- X_{t+1} \mid X_t = k)$ be the progress in one iteration. Recall $\eta := e/(e-1)$ and $q = \eta^{-\lambda}$.   Then
\begin{align*}
    \Prob(\Delta_k = 1) \ge\;& 1-\left(1 - \frac{k}{en}\right)^\lambda \ge \frac{\lambda k}{en + \lambda k} \ge \frac{1}{2}\min\left\{1,\frac{\lambda k}{en}\right\},\\
    \Prob(\Delta_k < 0) \le\;& q,\\
    \E(\Delta_k \mid \Delta_k < 0) \le\;& \eta.
\end{align*}
If $\lambda \ge 8nq/(k+1)$ and $\lambda = \omega(1)$,
\begin{align*}
    \E(\Delta_k) \ge \E(\min\{\Delta_k, 1\}) \ge\;& \min\left\{1, \frac{\lambda k}{n}\right\} \cdot \frac{3-e}{9}.
\end{align*}
\end{lemma}
}
\inLongVersion{
\begin{proof}
The first three statements are shown in~\cite[Lemma~2.2]{hevia2021self}, except for the last step on $\Prob(\Delta_k = 1)$, which holds since either $en + \lambda k \le 2en$ or $en + \lambda k \le 2\lambda k$.

We turn to a lower bound on $\E(\Delta_k)$.
The condition on~$\lambda$ implies $k+1 \ge 8n q/\lambda$ as in~\cite[Theorem~4.7]{Bossek2021a} (with $r \coloneqq k$ and $c \coloneqq 8 > e^2$ as required).
By the proof of said theorem, the drift of the \ocl in terms of the Hamming distance to the optimum, if the current Hamming distance is $k$, is at least
\[
\begin{cases}
    \frac{3-e}{6}, & \text{if $\lambda k \ge (3-e) n$,}\\
    \frac{\lambda k}{n} \cdot \frac{3-e}{9}, & \text{otherwise.}
\end{cases}
\]
Thus, the drift is at least $\min\{1, \frac{\lambda k}{n}\} \cdot \frac{3-e}{9}$ for all $k \ge 1$.
We note that the cited lower bounds on the drift, and the above arguments, only consider steps decreasing the distance by exactly~1.
Hence the drift bounds remain valid for the drift of $\min\{\Delta_k,1\}$.
\end{proof}
}
\inLongVersion{The next lemma confirms that the probability of not creating a clone is $\approx q$.}

\inShortVersion{\setcounter{theorem}{1}}
\begin{lemma}
\label{lem:probability-of-no-clone}
Let $\lambda = o(n)$ and $q =\eta^{-\lambda}$. Then the probability of an iteration of the \ocl and the \opl of not creating a clone of the current search point is $q(1+o(1))$.
\end{lemma}
\inLongVersion{
\begin{proof}
The probability of not creating a clone during $\lambda$ offspring creations is 
\[
    \left(1 - \left(1 - \tfrac{1}{n}\right)^n\right)^\lambda,
\]
as all $\lambda$ offspring creations must flip at least one bit, and the latter happens with probability $1-(1-1/n)^n$.
Using $(1-1/n)^n \le 1/e \le (1-1/n)^{n-1}$, we get
\[
    \left(1 - \left(1 - \tfrac{1}{n}\right)^n\right)^\lambda \ge \left(1 - \tfrac{1}{e}\right)^\lambda = q,
\]
and
\begin{align*}
    \left(1 - \left(1 - \tfrac{1}{n}\right)^n\right)^\lambda 
    =\;& \left(1 - \left(1 - \tfrac{1}{n}\right)^{n-1}\left(1 - \tfrac{1}{n}\right)\right)^\lambda\\
    \le\;& \left(1 - \tfrac{1}{e}\left(1 - \tfrac{1}{n}\right)\right)^\lambda\\
    =\;& \left(1 - \tfrac{1}{e}\right)^\lambda \left(1 + \tfrac{1}{(e-1)n}\right)^\lambda\\
    \le\;& q \cdot \exp\left(\tfrac{\lambda}{(e-1)n}\right) = q \cdot (1+o(1)).\qedhere
\end{align*}
\end{proof}
}
\inLongVersion{Next we show that it is unlikely to have any mutation which flips more than $c\ln n$ bits within $n^2$ generations. Note that the following lemma is concerned with the first $n^2\ln(n)$ function evaluations, but this translated into $\Omega(n^2)$ generations if $\lambda = O(\ln(n))$.}

\begin{lemma}\label{lem:hamming-log}
    For every constant $c > 0$, the probability of an offspring having a Hamming distance of at least $c\ln(n)$ to its parent is $n^{-\Omega(\ln \ln n)}$. Hence, w.h.p.\ each offspring generated during the first $n^2\ln(n)$ evaluations in \ocl or \opl has Hamming distance at most $c\ln(n)$ from its parent.
\end{lemma}
\inLongVersion{
\begin{proof}
    The probability that a standard bit mutation flips at least $c\ln(n)$ bits is at most
\[
\begin{aligned}
\binom{n}{c \ln(n)} n^{-c\ln(n)} \le \frac{1}{(c \ln(n))!} = (\ln(n))^{-\Omega(\ln(n))} = n^{-\Omega(\ln \ln(n))},
\end{aligned}
\] 
so the second part of the statement follows by a union bound over $n^2$ offspring.
\end{proof}
}

\subsection{Domination Results}\label{sec:domination}
We continue with domination results. 
It was first shown in~\cite{doerr2012multiplicative} that \onemax is the easiest function for the \oea, and these results were extended later in~\cite{Sudholt2012c,Witt2013,doerr2019analyzing,doerr2021runtime}. Here we show that \onemax is also the easiest function for the \opl, and that conversely the \opl is the fastest mutation-based algorithm on \onemax which creates solutions in batches, see Theorem~\ref{thm:couple-com-plus}. Both also hold in fixed-target settings. This result has very powerful implications, and we first give three immediate consequences in Theorem~\ref{thm:domination}. We say that a random variable $Y$ \emph{stochastically dominates} by a random variable~$X$ if $\Prob(Y\ge s) \ge \Prob(X\ge s)$ for all $s\in \R$.  

\begin{theorem}\label{thm:domination}
Let $a,b \in[0,n]$, $\lambda \in \N$, and consider the \oea, the \ocl and the \opl with the same mutation rate $r\le 1/2$ and with starting points $x^{\sss{\mathrm{one}}}$, $x^{\sss{\mathrm{com}}}$ and $x^{\sss{\mathrm{plus}}}$ respectively. Let $f:\{0,1\}^n\to \R$ be any fitness function.
\begin{enumerate}[(a)]
\item 
Assume $\OM(x^{\sss{\mathrm{com}}}) \le a$ and $\OM(x^{\sss{\mathrm{plus}}}) \ge a$. Let $T^{\sss{\plus}}(\OM)$ be the number of rounds until the \opl  on \onemax creates a search point $x$ with $\OM(x) \ge b$, and let $T^{\sss{\mathrm{com}}}(f)$ be the number of rounds until the \ocl on $f$ creates a search point $x$ with $\OM(x) \ge b$. 
%
    Then $T^{\sss{\mathrm{com}}}(f)$ stochastically dominates $T^{\sss{\plus}}(\OM)$. 
    %
    \item Assume $\OM(x^{\sss{\mathrm{one}}}) \ge a$ and $\OM(x^{\sss{\mathrm{plus}}}) \le a$. On \onemax, let $T^{\sss{\mathrm{one}}}$ and $T^{\sss{\mathrm{plus}}}$ be the number of function evaluations until the respective algorithm finds a search point $x$ with $\OM(x) \ge b$. Then $T^{\sss{\mathrm{plus}}}$ stochastically dominates $T^{\sss{\mathrm{one}}}$.
    \item Let $x^\ast\in\{0,1\}^n$ and let $x^{\sss{f}}, x^{\sss{\OM}} \in \{0,1\}^n$ be such that $H(x^{\sss{f}},x^\ast) \ge a$ and $H(x^{\sss{\OM}},\vec 1) \le a$. 
    Let $T^{\sss{f}}$ be the hitting time of the set $\{x: H(x,x^\ast) \le b\}$ for the \opl with $x^{\sss{\mathrm{plus}}} = x^{\sss{f}}$ on $f$ and let $T^{\sss{OM}}$ be the hitting time of the set $\{x: H(x,\vec 1) \le b\}$ for the \opl with $x^{\sss{\mathrm{plus}}} = x^{\sss{\OM}}$ on \onemax. Then $T^{\sss{f}}$ stochastically dominates $T^{\sss{\OM}}$. 
\end{enumerate} 
\end{theorem}    
%
Part (b) was already known~\cite{jansen2005choice}. For (c), the most natural case is that $x^\ast$ is the unique global optimum of $f$, but this is not required. 

In fact, all three parts of Theorem~\ref{thm:domination} are just special cases of the following, more general theorem. It says that the \opl on \onemax is faster than any other mutation-based algorithm with the same mutation rate if it creates offspring in batches of size $\lambda$. Crucially, this holds for any selection strategy for the parents. Thus, it is also independent of the fitness function, since this ``only'' decides which individuals may reproduce.

\begin{theorem}\label{thm:couple-com-plus}~
Let $\lambda \in\N$, $a,b \in [0,n]$, mutation rate $r\le 1/2$ and let $x^{\sss{\CA}},x^{\sss{\mathrm{plus}}} \in \{0,1\}^n$ with $\OM(x^{\sss{\CA}}) \le a$ and $\OM(x^{\sss{\mathrm{plus}}})\ge a$. Consider any algorithm $\CA$ with the following scheme. The algorithm starts by creating $x^{\sss{\CA}}$. In each round, it uses an arbitrary mechanism to select $\lambda$ (not necessarily distinct) parents among all previously created search points, and creates $\lambda$ offspring by applying standard bit mutation with mutation probability $r$ to them.

    Let $S_t$ be the set of search points that $\CA$ creates in the first $t$ rounds, and let $X^{\sss{t,\CA}} := \max\{\OM(x) \mid x\in S_t\}$. Let $X^{\sss{t,\plus}}$ be the \OM-value of the \opl with standard bit mutation and mutation probability $p$ on \onemax after $t$ rounds if started in $x^{\sss{\mathrm{plus}}}$. Then $X^{\sss{t,\plus}}$ stochastically dominates $X^{\sss{t,\CA}}$.

    Moreover, let $T^{\sss{\CA}} := \min\{t: X^{\sss{t,\CA}} \ge b\}$ and $T^{\sss{\plus}} := \min\{t: X^{\sss{t,\plus}} \ge b\}$. Then $T^{\sss{\CA}}$ stochastically dominates $T^{\sss{\plus}}$.
\end{theorem}
\inShortVersion{The proofs are based on Lemma 6.1 in~\cite{Witt2013}, omitted due to space limitations. We remark that we could have replaced the \ocl in Theorem~\ref{thm:domination}(a) by any other algorithm as in Theorem~\ref{thm:couple-com-plus}.}
\inLongVersion{
\begin{proof}
The proof is based on Lemma 6.1 in~\cite{Witt2013}. This says that if $y$ and $y'$ are obtained by standard bit mutation of $x$ and $x'$ respectively, with mutation probability $r\le 1/2$, and if $\OM(x)\le \OM(x')$ then $\OM(y')$ stochastically dominates $\OM(y)$.  

We show that for all $b\in [0,n]$,
    \begin{align}\label{eq:domination-rounds}
        \Prob(X^{\sss{t,\CA}} \ge b) \le \Prob(X^{\sss{t,\plus}} \ge b).
    \end{align}
We use induction over $t$. For $t=0$, \eqref{eq:domination-rounds} is satisfied because the \opl starts with 
$\OM(x^{\sss{\mathrm{plus}}}) \ge a$, while $\CA$ starts with $\OM(x^{\sss{\CA}}) \le a$. So we assume that~\eqref{eq:domination-rounds} holds for some $t \ge 0$ and show the same statement for $t+1$. Since $X^{\sss{t,\plus}}$ stochastically dominates $X^{\sss{t,\CA}}$, we can couple them such that $X^{\sss{t,\plus}}\ge X^{\sss{t,\CA}}$. 
We will show that for all $s\in [1,n]$, 
\begin{align}\label{eq:domination-rounds-2}
\Prob\big([X^{\sss{t+1,\CA}} \ge b \mid X^{\sss{t,\plus}} = s\big) \le \Prob\big(X^{\sss{t+1,\plus}} \ge b \mid X^{\sss{t,\plus}} = s\big).
\end{align}
Note that we condition on the same event on both sides, so~\eqref{eq:domination-rounds} follows from~\eqref{eq:domination-rounds-2} due to
\begin{align*}
\Prob\big(X^{\sss{t+1,\CA}} \ge b\big) & = \sum_{s} \Prob\big(X^{\sss{t,\plus}} = s\big)\Prob\big(X^{\sss{t+1,\CA}} \ge b \mid X^{\sss{t,\plus}} = s\big) \\
& \le \sum_{s} \Prob\big(X^{\sss{t,\plus}} = s\big)\Prob\big(X^{\sss{t+1,\plus}} \ge b \mid X^{\sss{t,\plus}} = s\big)\\
&= \Prob\big(X^{\sss{t+1,\plus}} \ge b\big).
\end{align*}

In the case $s\ge b$, there is nothing to show since the \opl is elitist and thus the right-hand side of~\eqref{eq:domination-rounds-2} is one. So let us fix some value $s <b$ such that $X^{\sss{t,\plus}} = s$. Denote by $p_{\text{imp}} = p_{\text{imp}}(s)$ the probability that the first offspring of the \opl in round $t+1$ creates an offspring of $\OM$-value at least $b$. Then $\Prob\big(X^{\sss{t+1,\plus}} \ge b \mid X^{\sss{t,\plus}} = s\big) = 1-(1-p_{\text{imp}})^\lambda$. On the other hand, since $X^{\sss{t,\CA}} \le  X^{\sss{t,\plus}} = s$, the algorithm $\CA$ has only created potential parents of $\OM$-value at most $s$ until round $t$. Hence, by~\cite[Lemma 6.1]{Witt2013}, any offspring $y$ of $\CA$ in round $t+1$ satisfies $\Prob\big(\OM(y) \ge b \mid X^{\sss{t,\plus}} = s\big) \le p_{\text{imp}}$, and this bound holds independently for all $\lambda$ offspring of $\CA$. Hence,
\begin{align*}
    \Prob\big(X^{\sss{t+1,\CA}} \ge b \mid X^{\sss{t,\plus}} = s\big) & \le 1-(1-p_{\text{imp}})^\lambda \\
    & = \Prob\big(X^{\sss{t,\plus}} \ge b \mid X^{\sss{t+1,\plus}} = s\big).
\end{align*}
This concludes the induction and proves the first domination statement. The second domination statement is just a reformulation of the first one: the event ``$T^{\sss{\CA}} \le t$'' is identical to the event ``$X^{\sss{t,\CA}} \ge b$'', since both express that within the first $t$ rounds $\CA$ creates a search point of $\OM$-value at least $b$. Hence,
    \begin{equation*}
    \Prob\big(T^{\sss{\CA}} \le t\big)  = \Prob\big(X^{\sss{t,\CA}} \ge b\big)  \le \Prob\big(X^{\sss{t,\plus}} \ge b\big)   = \Prob\big(T^{\sss{\plus}} \le t\big).\qedhere
    \end{equation*} 
\end{proof}
}
\inLongVersion{
\begin{proof}[Proof of Theorem~\ref{thm:domination}]
    For part (a), we just need to observe that the \ocl on any fitness function $f$ falls into the category of $\CA$ in Theorem~\ref{thm:couple-com-plus}. So we may choose the \ocl for $\CA$. 

    Part (b) follows by setting $\lambda_{\ref{thm:couple-com-plus}} := 1$ in Theorem~\ref{thm:couple-com-plus}, where we use the subscript $\ref{thm:couple-com-plus}$ to distinguish it from the $\lambda$ in the statement of this corollary. Then the $(1+\lambda_{\ref{thm:couple-com-plus}})$ EA from that lemma is just the \oea, and for algorithm $\CA$ we can take the \opl. 
    Then Theorem~\ref{thm:couple-com-plus} implies that $T^{\sss{\plus}}$ stochastically dominates $T^{\sss{\mathrm{one}}}$.

    For (c), by symmetry it suffices to show the statement for the case $x^\ast = \vec 1$. We again apply Theorem~\ref{thm:couple-com-plus}, where we choose for $\CA$ the \opl on $f$. 
    Then 
    Theorem~\ref{thm:couple-com-plus} implies that $T^{\sss{f}}$ stochastically dominates $T^{\sss{\OM}}$.
\end{proof}

As the proof shows, we could have replaced the \ocl in Theorem~\ref{thm:domination}(a) by any other algorithm as in Theorem~\ref{thm:couple-com-plus}.
}


\subsection{High-Probability Fixed Target Results}\label{sec:fixed-target}

Now we give upper and lower bounds for the time that the \opl and the \ocl need to reach some target fitness on \onemax, in the regime where the \ocl is efficient. We show that it is exponentially unlikely to deviate from the expectation by more than a constant factor. 

The time bounds match known ones for the \opl~\cite{Lassig2011,GiessenW17} and the \ocl~\cite{jagerskupper2007plus,rowe2014choice,hevia2021self,Bossek2021a}, albeit that the dependency on~$\lambda$ can be improved slightly, see~\cite{Doerr2013,Badkobeh2014,Lehre2019} and we do not have tight leading constants. The strength of our result lies in its generality and exponentially small tail bounds.
Related previous work includes upper tail bounds~\cite{DoerrG13} and lower tail bounds for \onemax%
~\cite{LehreWitt2021},
a review of fixed-target results in~\cite{Buzdalov2022_fixedtarget} and
black-box complexity lower bounds with tail bounds for unary unbiased black-box algorithms~\cite{Lehre2019} in a framework similar to ours. The latter work includes a fixed-target scenario of getting close (in Hamming distance) to global or local optima. Our tail bounds are stronger than the previous ones.

\begin{theorem}
\label{the:big-ass-fixed-target-theorem}
Consider an algorithm $\CA$ as in Theorem \ref{thm:couple-com-plus} with $r \le 1/2$ and a fitness function $f \colon \{0, 1\}^n \to \R$.
Let $a, b \in [0,n]$ with $a > b$ and fix a search point $x^\ast\in \{0, 1\}^n$.
Let $T^{\sss{\CA, f}}$ denote the number of evaluations made by $\CA$ on $f$, starting with a population of search points that all have Hamming distance at least $a$ to~$x^\ast$, to reach a search point within Hamming distance at most~$b$ of $x^\ast$.

There are positive constants $c_1, c_2$ such that the following holds.
\begin{enumerate}
    \item For every algorithm $\CA$, every fitness function~$f$ and every target search point $x^\ast$,
    \[
        \Prob\left(T^{\sss{\CA, f}} \le c_1 n\ln(a/b)\right) \le e^{-\Omega(\min\{a-b, b\})}.
    \]
    \item For $f = \onemax$, $x^\ast = \vec{1}$ and either $\CA = $\;\opl with arbitrary $\lambda$ (including the (1+1)~EA) or $\CA = $\;\ocl with $\lambda \ge \max\{\loge(n/b), 16(e+1)/3\}$, 
    \[
        \hspace*{1cm} \Prob\left(T^{\sss{\CA, f}} \ge c_2\left((a-b)\lambda + \ln(a/b)n\right)\right) \le e^{-\Omega(\min\{a-b, b\})}.
    \]
\end{enumerate}
\end{theorem}

\inShortVersion{
We omit the proof of Theorem~\ref{the:big-ass-fixed-target-theorem}. It is not based on drift analysis. Instead, one can prove lower tail bounds for the (1+1)~EA by arguing that in every step at most $a$ 0-bits may flip and then applying Chernoff bounds to bound the total number of flipping 0-bits, and hence the possible distance decrease, in the stated time. This approach works well if $b = \Omega(a)$. Otherwise, we consider $\approx \log_2(a/b)$ distance intervals $[a\cdot 2^{-1}, a], [a \cdot 2^{-2}, a \cdot 2^{-1}], \dots$ and apply the above arguments for each interval, always considering the worst case for the number of ones. By Theorem~\ref{thm:domination}, the lower tail bound for the (1+1)~EA translates to the stated general setting. 

Upper bounds are derived using a similar division into intervals, but we have to account for non-elitism. We use Chernoff bounds to obtain a lower bound for the number of steps in which the distance to the optimum is changed (by positive or negative values). Then we show that, irrespective of the current fitness, the conditional expected progress in distance-changing steps stochastically dominates a random variable with exponentially decaying tail:
\begin{align}
    Z_i \coloneqq\;&
\begin{cases}
    +1 & \text{with probability~$\frac{3}{4}$}\\
    -j & \text{with probability~$\frac{3}{4} \cdot 4^{-j}$, for $j \in \mathbb{N}$}.
\end{cases}\label{eq:simplified-progress-sketch}
\end{align}
We prove a Chernoff-type bound for the sum $Z \coloneqq \sum_{i=1}^t Z_i$ and show
$\Prob\left(Z \le \E(Z)/2\right) \le e^{-7t/8}$. Together, this bounds the total progress in each interval, with the claimed tail bounds.
}

\inLongVersion{
We start with the upper tail bound. Note that it suffices to prove the tail bound for the \oea on \onemax, since by Theorem~\ref{thm:domination} the same tail bounds on the number of function evaluations also hold for the \opl on \onemax by (b) and for any other $\CA$ and $f$ by Theorem~\ref{thm:couple-com-plus}.

\begin{lemma}
\label{lem:simple-lower-bound-with-Chernoff}
Let $a, b \in [n]$ with $a > b$. Let $T$ denote the number of iterations for the (1+1)~EA to reach a distance of at most~$b$ from $\vec 1$ when starting at distance~$a$ on \onemax. Then 
\[
    \Prob\left(T \le \frac{(a-b)n}{2a}\right) \le e^{-(a-b)/6}.
\]
Moreover, if $a \ge 2b$ then
\[
    \Prob\left(T \le \frac{\log_2(a/b)n}{16}\right) \le e^{-\Omega(b)}.
\]
\end{lemma}
\begin{proof}
In order to go from distance $a$ to $b$ in at most $(a-b)n/2a$ rounds, there must be at least $a-b$ bit flips among the initial $a$ positions of $0$-bits during those rounds.
In each round, each of the $a$ positions has probability $1/n$ to be flipped. In $(a-b) \cdot \frac{n}{2a}$ rounds, there are thus $a\cdot (a-b) \cdot \frac{n}{2a} = (a-b)n/2$ chances for these bits to flip, each with probability $1/n$. Thus, the total number of bit flips among the $a$ positions is given by a Binomial distribution $\Bin((a-b)n/2,1/n)$, which has expectation $(a-b)/2$. By Chernoff bounds, the probability of having at least $a-b$ bit flips among the $a$ positions is at most
\[
    \exp\left(-\frac{(a-b)}{6}\right).
\]


For the second statement, we apply the first statement 
repeatedly. Let $\ell$ be the smallest integer such that $a/2^\ell < b$. 
Since $a \ge 2b$, we have $\ell > 1$.
Now we apply the first statement with parameters $a_1, b_1$ chosen as $a_1 \coloneqq a$ and $b_1 \coloneqq a_1/2$.
At the end of the considered time period, a distance of at most $b_1$ is reached. Note that the distance may be smaller than $b_1$ in case the last jump overshoots the target of $b_1$. However, this jump decreases the distance by at most $b_1/4$, with probability $1 - \exp(-\Omega(b_1 \ln b_1))$ since at least $b_1/4$ bits would have to flip. So we may assume that this happens. Thus, we can iterate the argument and apply the first statement 
 with $a_2$ chosen as the distance reached when it decreases below~$b_1$ for the first time, and $b_2 \coloneqq a/2^2$. Note that $a_2 - b_2 \ge a/2^2$. We iterate these arguments $\ell-1$ times, choosing $b_{\ell-1} \coloneqq b$.

Since the interval in the $i$th application of the first statement is $(a_i - b_i) \ge a/2^{i+1}$ and $a_i \le a/2^{i-1}$, we get a lower bound of
\[
    \sum_{i=1}^{\ell-1} \frac{(a_i-b_i)n}{2 a_i}
    \ge \sum_{i=1}^{\ell-1} \frac{an/2^{i+1}}{2 a/2^{i-1}}
    = \frac{n}{8} \sum_{i=1}^{\ell-1} 1
    \ge \frac{\ell n}{16}
\]
using $\ell \ge 2$ in the last step.
Since $a/2^{\ell} < b$ and $\ell > \log_2(a/b)$, this matches the claimed time bound. Taking a union bound over all failure probabilities yields a total failure probability of at most
\[
    \sum_{i=1}^{\ell-1} \left(e^{-a/(6\cdot 2^{i+1})} + e^{-\Omega(a/2^i \ln(a/2^i))}\right) = e^{-\Omega(b)}. \qedhere
\]
\end{proof}




For the lower tail bound, we couple the progress to the following set of independent random variables.

\begin{definition}
\label{def:simplified-progress-variables}
For $i \in \{1, \dots, t\}$ 
define independent random variables
\begin{align*}
    Z_i \coloneqq\;&
\begin{cases}
    +1 & \text{with probability~$\frac{3}{4}$}\\
    -j & \text{with probability~$\frac{3}{4} \cdot 4^{-j}$, for $j \in \mathbb{N}$}.
\end{cases}
\end{align*}
Note that the $Z_i$ are iid with expectation
\[
\E(Z_1) = \frac{3}{4} - \sum_{j=1}^\infty \frac{3}{4} \cdot j4^{-j} = \frac{3}{4} \left(1 - \frac{4}{9}\right) = \frac{5}{12}, 
\]
where we used the equality $\sum_{j=1}^\infty j x^j = \frac{x}{(1-x)^2}$ for $0 < x < 1$.
%
%

\end{definition}

\begin{lemma}
\label{lem:progress-dominates-Z}
Consider the \ocl on \onemax with $k^\ast \le n/\lambda$ and $\lambda \ge \max\{\loge(n/k^\ast), 16(e+1)/3\}$. Then for all~$k \ge k^\ast$ the progress $(X_t - X_{t+1} \mid X_t = k, X_{t+1} \neq X_t)$, conditional on $X_{t+1} \neq X_t$, stochastically dominates $Z_i$.
\end{lemma}
\begin{proof}
Let $\Delta_k\!\coloneqq\! (X_t - X_{t+1} \mid X_t = k)$ and $C\! \coloneqq\! \tfrac{16}{3}$.
By Lemma~\ref{lem:transition-probabilities},
\[
    \Prob(\Delta_k \ge 1) \ge \frac{\lambda k}{en + \lambda k},
\]
and
\[
    \Prob(\Delta_k < 0) \le q = \eta^{-\lambda} \le \frac{k^\ast}{n},
\]
where we have used the precondition on $\lambda$. 
Using the assumption $\lambda \ge C(e+1)$ and recalling $k^\ast \le n/\lambda$,
\[
    \frac{k^\ast}{n} \le \frac{1}{C} \cdot \frac{\lambda k^\ast}{en+n} \le \frac{1}{C} \cdot \frac{\lambda k^\ast}{en + \lambda k^\ast} \le \frac{1}{C} \cdot \frac{\lambda k}{en + \lambda k} \le \frac{1}{C} \cdot \Prob(\Delta_k \ge 1).
\]
Thus, we get
\begin{equation}
\label{eq:delta-k-negative-one}
    \Prob(\Delta_k = -1) \le \Prob(\Delta_k < 0) \le  \frac{1}{C} \cdot \Prob(\Delta_k \ge 1).
\end{equation}
Now fix $j\in[2, n-k]$. The \ocl only increases its distance to $\vec 1$ by~$j$ if all $\lambda$ offspring increase the distance by at least~$j$.
A necessary condition is that
at least $j$ bits flip in all $\lambda$ offspring. The probability of this event is at most
\begin{align*}
    \Prob(\Delta_k = -j) \le \left(\binom{n}{j} n^{-j}\right)^\lambda
    \le\;& \left(\frac{1}{j!}\right)^\lambda =
    \eta^{-\lambda} \left(\frac{\eta}{j!}\right)^\lambda.
\end{align*}
The term $q= \eta^{-\lambda}$ was already bounded by $\frac{1}{C} \cdot \Prob(\Delta_k \ge 1)$ in~\eqref{eq:delta-k-negative-one}. The term $ (\eta/j!)^{\lambda}$ is at most $C^{-j+1}$ for all $j \ge 2$ if $\lambda \ge 6$. Thus, we have shown $\Prob(\Delta_k = -j) \le C^{-j} \cdot \Prob(\Delta_k \ge 1)$. This implies
\[
    \Prob(\Delta_k \neq 0) \le \Prob(\Delta_k \ge 1) + \sum_{j=1}^\infty C^{-j} \cdot \Prob(\Delta_k \ge 1) = \frac{C}{C-1} \cdot \Prob(\Delta_k \ge 1).
\]
Now the claim follows from
\begin{align*}
    & \Prob(\Delta_k \ge 1 \mid \Delta_k \neq 0)\\
    & = \frac{\Prob(\Delta_k \ge 1)}{\Prob(\Delta_k \neq 0)} \ge \frac{\Prob(\Delta_k \ge 1)}{\frac{C}{C-1} \cdot \Prob(\Delta_k \ge 1)} = \frac{C-1}{C} \ge \frac{3}{4} = \Prob(Z_i = 1)
\end{align*}
and, for all $j \in \mathbb{N}$,
\begin{align*}
     & \Prob(\Delta_k = -j \mid \Delta_k \neq 0)\\
     & = \frac{\Prob(\Delta_k = -j)}{\Prob(\Delta_k \neq 0)} \le \frac{C^{-j} \cdot \Prob(\Delta_k \ge 1)}{\Prob(\Delta_k \ge 1)} 
    \le \frac{3}{4} \cdot 4^{-j} = \Prob(Z_i = -j). \qedhere
\end{align*}
\end{proof}

Now we give a Chernoff-type deviation bound for the sum of $Z_i$ variables.
\begin{lemma}
\label{lem:Chernoff-type-bounds-for-Z}
Consider random variables $Z_1, \dots, Z_t$ and $Z \coloneqq \sum_{i=1}^t Z_i$ as in Definition~\ref{def:simplified-progress-variables}. Then 
\[
    \Prob\left(Z \le \frac{\E(Z)}{2}\right) \le e^{-7t/8}.
\]
\end{lemma}
\begin{proof}
We follow the proof of Chernoff bounds. Assume $\gamma > 0$ is a constant chosen later, such that $e^\gamma/4 < 1$.
Using Markov's inequality and $\E(Z) = t \cdot \frac{5}{12}$,
\begin{align*}
    \Prob\left(Z \le \frac{\E(Z)}{2}\right)
    = \Prob(e^{-\gamma Z} \ge e^{-\gamma t \cdot 5/24})
    \le \frac{\E(e^{-\gamma Z})}{e^{-\gamma t \cdot 5/24}}.
\end{align*}
We simplify the numerator as follows, exploiting the independence of the $Z_i$'s:
\begin{align*}
    \E(e^{-\gamma Z}) = \E\left(\prod_{i=1}^t e^{-\gamma Z_i}\right)
    = \prod_{i=1}^t \E\left(e^{-\gamma Z_i}\right).
\end{align*}
By the density of the random variables $Z_i$ from Definition \ref{def:simplified-progress-variables},
\begin{align*}
    \E\left(e^{-\gamma Z_i}\right)
    =\;& \frac{3}{4} \cdot e^{-\gamma} + \sum_{j=1}^\infty e^{\gamma j} \cdot \frac{3}{4} \cdot 4^{-j}\\
    =\;& \frac{3}{4} \left(e^{-\gamma} + \frac{e^\gamma/4}{1-e^\gamma/4}\right).
\end{align*}
Choosing $\gamma \coloneqq \ln(4/3)$, this simplifies to
\[
    \E\left(e^{-\gamma Z_i}\right)=\frac{3}{4} \left(\frac{3}{4} + \frac{1/3}{1-1/3}\right) = \frac{15}{16}.
\]
Plugging this back in yields
\[
    \E(e^{-\gamma Z}) = \exp\left(- 15t/16\right),
\]
so
\[
   \frac{\E(e^{-\gamma Z})}{e^{-\gamma t \cdot 5/24}} = \exp\left(- t \cdot \frac{15}{16} + \gamma t \cdot \frac{5}{24}\right).
\]
Noting that $15/16 - \ln(4/3) \cdot 5/24 > 7/8$ completes the proof.
%
\end{proof}

The coupling allows us to derive tail bounds for the \ocl and the \opl on \onemax.

\begin{lemma}
\label{lem:upper-tail-bound-a-to-b}
Assume the conditions of Lemma~\ref{lem:progress-dominates-Z}. Let $a, b \in [n]$ with $a > b$. Let $T$ denote the random number of iterations for the \ocl or the \opl to reach a distance of at most~$b$ from $\vec 1$ when starting at distance~$a$ on \onemax. Then 
\[
    \Prob\left(T \ge (a-b) \cdot \frac{48}{5} \left(1 + \frac{en}{\lambda b}\right)\right) \le 2e^{-(a-b) \cdot 6/5}.
\]
Moreover, if $a \ge 2b$ then
\[
    \Prob\left(T \ge \frac{48}{5} \cdot (a-b) + \frac{48en}{5\lambda} \cdot (\log_2(a/b) + 1)\right) \le e^{-\Omega(b)}.
\]
\end{lemma}
\begin{proof}
We first consider the \ocl. 
Let $X_t$ denote the distance of the current search point to the optimum at time~$t$. Since we are only interested in reaching a distance of at most~$b$, we may assume that the process remains at distance~$b$ as soon as a distance of $\le b$ is reached for the first time.
We call a step~$t$ \emph{relevant} if $X_{t+1} \neq X_t$. 
As long as $X_t \ge b$, the probability of a relevant step is at least $\Prob(X_{t+1} > X_t \mid X_t = b) \ge \frac{\lambda b}{en + \lambda b} \eqqcolon p_b$ by Lemma~\ref{lem:transition-probabilities}. Let $r \coloneqq (a-b) \cdot \frac{24}{5}$ and $T \coloneqq 2r/p_b$. Thus, the number of relevant steps in $T$ iterations stochastically dominates a sum of $T$ iid Bernoulli random variables $Y_1, \dots, Y_T$ with parameters $p_b$.

Let $Y \coloneqq Y_1, \dots, Y_T$, then $\E(Y) = T p_b = 2r$. By Chernoff bounds, $\Prob(Y \le \E(Y)/2) = e^{-r/4} = e^{-(a-b)\cdot 6/5}$. By stochastic domination, this also constitutes an upper bound on the probability of having fewer than $r$ relevant steps.

Now assume the algorithm makes at least $r$ relevant steps. By Lemma~\ref{lem:progress-dominates-Z}, the total progress in these steps stochastically dominates the sum of $r$ variables, $Z \coloneqq Z_1 + Z_2 + \dots + Z_r$ where the $Z_i$ are defined as in Definition~\ref{def:simplified-progress-variables}. The expected progress is $\E(Z) = r \cdot \E(Z_1) = r \cdot \frac{5}{12}$. By Lemma~\ref{lem:Chernoff-type-bounds-for-Z}, the probability of the progress being at most $\E(Z)/2 = r \cdot \frac{5}{24} = a-b$ is at most $e^{-7r/8} = e^{-(a-b) \cdot 21/5}$.
Taking a union bound over the two failure events proves the first claim.

For the second statement, we apply the first statement 
repeatedly. Let $\ell$ be the smallest integer such that $a/2^\ell < b$. 
Since $a \ge 2b$, we have $\ell > 1$.
Now we apply the first statement $\ell-1$ times, with parameters $(a_1, b_1), (a_2, b_2), \dots, (a_{\ell-1}, b_{\ell-1})$ 
chosen as $a_1 \coloneqq a$, $b_{\ell-1} \coloneqq b$, $a_i \coloneqq a/2^{i-1}$ for $i \in [2, \dots, \ell-1]$ and $b_i \coloneqq a/2^i$ for $i \in [1, \dots, \ell-2]$.
Note that $a_i - b_i = a/2^{i}$ for all $i \in [1, \ell-2]$ and $a_{\ell-1} - b_{\ell-1} \in [a/2^{\ell-1}, a/2^{\ell-2}]$.

We get an upper bound bound of 
\begin{align*}
    & \sum_{i=1}^{\ell-1} (a_i-b_i) \cdot \frac{48}{5} \left(1 + \frac{en}{\lambda b_i}\right)\\
    =\;& \frac{48}{5} \sum_{i=1}^{\ell-1} (a_i-b_i) + \frac{48en}{5\lambda} \sum_{i=1}^{\ell-1} (a_i-b_i)\cdot \frac{1}{b_i}.
\end{align*}
The first sum is telescopic and simplifies to $a-b$. In the second sum, summands for $i \le \ell-2$ simplify to $1$ since $a_i-b_i = a/2^i = b_i$. The last summand is at most twice as large as the previous ones, hence the whole sum is at most $\ell$. By choice of~$\ell$, $a/2^{\ell-1} \ge b$ or, equivalently, $\ell \le \log_2(a/b) + 1$.
Together, the upper bound is
\[
    \frac{48}{5} \cdot (a-b) + \frac{48en}{5\lambda} \cdot (\log_2(a/b) + 1).
\]
Taking a union bound over all failure probabilities yields a total failure probability of at most
\[
    \sum_{i=1}^{\ell-1} 2e^{-a/2^{i} \cdot 6/5} = e^{-\Omega(b)}. 
\]
This completes the proof for the \ocl. The statement for the \opl follows since by Theorem~\ref{thm:domination} (a) the time $T$ for the \opl is dominated by the corresponding time for the \ocl.
\end{proof}

Now we can finally prove Theorem~\ref{the:big-ass-fixed-target-theorem}.
\begin{proof}[Proof of Theorem~\ref{the:big-ass-fixed-target-theorem}]
For the first statement, by symmetry we may assume $x^\ast =\vec 1$. By Theorem~\ref{thm:couple-com-plus}, the time $T^{\sss{\CA, f}}$ stochastically dominates $T^{\sss{\plus}}(\OM)$ and by Theorem~\ref{thm:domination} (b), the latter time stochastically dominates $T^{\sss{\mathrm{one}}}$ using the notation from Theorem~\ref{thm:domination}. 
Hence it suffices to prove the statement for $\CA =$\;(1+1)~EA and $f=\onemax$.

We choose $c_1 \coloneqq 1/(16\ln(2))$. If $b \ge a/2$ then we argue as follows.
Using $x \ge \ln(1+x)$ for all $x \in \R$, we get
\[
    \frac{a-b}{2a} \ge \frac{1}{4} \cdot \frac{a-b}{b} \ge \frac{1}{4} \cdot \ln\left(1 + \frac{a-b}{b}\right) = \frac{1}{4} \cdot \ln(a/b) \ge c_1 \ln(a/b).
\]
Along with Lemma~\ref{lem:simple-lower-bound-with-Chernoff},
\[
    \Prob\left(T^{\sss{\mathrm{one}}} \le c_1 \ln(a/b)n \right) \le \Prob\left(T^{\sss{\mathrm{one}}} \le \frac{(a-b)n}{2a}\right) \le e^{-(a-b)/6}.
\]
As the exponent is $-\Omega(\min\{a-b, b\})$, this implies the claim.

If $b < a/2$ then Lemma~\ref{lem:simple-lower-bound-with-Chernoff} yields
\[
    \Prob\left(T^{\sss{\mathrm{one}}} \le \frac{\log_2(a/b)n}{16}\right) \le e^{-\Omega(b)} = e^{-\Omega(\min\{a-b, b\})}
\]
and this implies the claim since $\log_2(a/b)/16 = c_1 \ln(a/b)$.

For the second statement we aim to apply Lemma~\ref{lem:upper-tail-bound-a-to-b}. We must choose $k^\ast \le n/\lambda$ such that the conditions on~$\lambda$ are met, that is, such that $\lambda \ge \max\{\loge(n/k^\ast), 16(e+1)/3\}$. Recall that we may use the condition $\lambda \ge \max\{\loge(n/b), 16(e+1)/3\}$. 
Since this implies $\lambda \ge 16(e+1)/3$, we focus on the respective first arguments of the $\max$ terms. 
If $b \le n/\lambda$, we take $k^\ast \coloneqq b$ in Lemma~\ref{lem:upper-tail-bound-a-to-b} and
note that  $\loge(n/b) \ge \loge(n/k^\ast)$. 
If $b > n/\lambda$, we take $k^\ast \coloneqq n/\lambda$ and the condition from Lemma~\ref{lem:upper-tail-bound-a-to-b} simplifies to 
$\lambda \ge \loge(\lambda)$, which is true for all $\lambda \in \N$. Thus, Lemma~\ref{lem:upper-tail-bound-a-to-b} is applicable.

Now we choose $c_2 \coloneqq \frac{96e}{5\ln 2}$.
If $b \ge a/2$ then we apply $\frac{x}{x+1} \le \ln(1+x)$ for all $x > -1$ to $x \coloneqq (a-b)/b$ and get
\[
    \frac{a-b}{b} \le 2\cdot  \frac{a-b}{a} = 2 \cdot \frac{(a-b)/b}{a/b} \le 2\ln(1+(a-b)/b) = 2\ln(a/b).
\]
Note that Lemma~\ref{lem:upper-tail-bound-a-to-b} bounds the number of iterations, hence we need to multiply with~$\lambda$ to obtain a bound on the number of function evaluations. This yields an upper bound on the number of evaluations of
\begin{align*}
    \frac{48}{5} \cdot (a-b)\lambda + \frac{48en}{5} \cdot \frac{a-b}{b}
    \le \frac{48}{5} \cdot (a-b)\lambda + \frac{96en}{5} \cdot \ln(a/b)
\end{align*}
that holds with probability $1-2e^{-(a-b) \cdot 6/5} \ge 1-e^{-\Omega(a-b)}$.
Since $48/5 \le 96e/5 \le c_2$, this proves the claim.

If $b < a/2$ then we note that
\[
    \log_2(a/b)+1 \le 2\log_2(a/b) = \frac{2}{\ln 2} \cdot \ln(a/b).
\]
Lemma~\ref{lem:upper-tail-bound-a-to-b}, multiplied by~$\lambda$, yields a bound of 
\[
    \frac{48}{5} \cdot (a-b)\lambda + \frac{48en}{5} \cdot (\log_2(a/b)+1) 
    \le \frac{48}{5} \cdot (a-b)\lambda + \frac{96en}{5\ln 2} \cdot \ln(a/b).
\]
Noting that both leading constants are at most $c_2$ and that the failure probability from Lemma~\ref{lem:upper-tail-bound-a-to-b} is $e^{-\Omega(b)} = e^{-\Omega(\min\{a-b, b\})}$ completes the proof.
\end{proof}
}


\section{Comma strategy escapes traps}\label{sec:com-upper}
In this section we prove the upper bound on $\Tcom$ in~\eqref{eq:maincom} in Theorem~\ref{thm:main}. 
The proof consists of three steps: first we introduce \dydisOM, a ``dynamic'' version of \disOM, and study the drift of the \ocl on \dydisOM using drift analysis for a suitable potential function. Studying  \dydisOM instead of \disOM simplifies the drift analysis, since the ``frozen'' noise in \disOM introduces dependencies with respect to distorted points that are hard to control. Using the bounds on the drift, we compute the expected running time of the \ocl on \dydisOM. Unfortunately, our potential does not satisfy the conditions to employ a standard additive drift theorem with tail bounds. Still, we obtain an ``almost matching'' bound on the running time which holds w.h.p. 

Using the non-sharp upper bound, we develop a `ratchet' argument (cf.~\cite{hevia2021self}): we show that the algorithm never moves more than $o(\ln^4(n))$ away from the target. This event we use to sharpen our upper bound to obtain concentration around the expected running time. This bootstrapping argument requires a fine control on the drift. In particular, we need to consider steps towards the optimum by more than $1$ (contrary to the process in \inLongVersion{ Definition~\ref{def:simplified-progress-variables}}%
\inShortVersion{\eqref{eq:simplified-progress-sketch}}).

Moreover, the ratchet argument allows to argue that no distorted search point is evaluated twice. This will imply that w.h.p.\ $\Tcomdyn=\Tcom$, i.e., the number of function evaluations until the \ocl finds a search point of fitness at least $n-k^\ast$ on \dydisOM, is the same as the number of function evaluations on \disOM. 

\paragraph{\textsc{Dynamic distorted OneMax.}} We introduce a dynamic version of \disOM in which we reveal the sets of distorted and clean points gradually, and in which previously sampled distorted points can become clean later (but not the other way around). Let $s:=t\lambda+j$ for $j\in[\lambda]$, write $y^\sss{(s)}$ for the $s$th sampled search point after initialization, and $x^\sss{(t)}$ for the current search point. With $\CC_0=\{x^\sss{(0)}\}$, we iteratively define for $s\ge 1$
\[
\CC_{s} = 
\begin{dcases}
\CC_{s-1}\cup\{y^\sss{(s)}\},&\text{w/p }1-p, \text{if }y^\sss{(s)}\neq x^\sss{(t)},\\
\CC_{s-1},&\text{otherwise.}
\end{dcases}
\]
Note that $\CC_{s-1}\subseteq\CC_{s}$ for all $s$, reflecting that clean points remain clean forever.
Given $\CC_{s}$, we define
\[
\dydisOM(y^\sss{(s)}):=
\begin{dcases}
    \OM(x),&\text{if }x\in\CC_s, \\
    \OM(x)+d,&\text{otherwise}.
\end{dcases}
\]

We use drift analysis~\cite{lengler2020drift} to analyse the \ocl on \dydisOM. For convenience, we use a potential that decreases with the distance from the target, so we use the \emph{\zeromax} function $\ZM(x) := n-\OM(x)$. We frequently abbreviate $\ZM(x)=k$. We introduce some extra notation to define the potential function. Let $Y_1,\ldots,Y_\lambda\sim\mathrm{Bin}(k, 1/n)$ be iid binomial random variables, and define $Y^\ast(k):=\max(Y_1,\ldots,Y_\lambda)$. The random variable $Y_i$ represents the number of $0$-bits flipped into a $1$ by the $i$th offspring when the parent has $\ZM(x)=k$. 
\inLongVersion{Next we define a potential function which penalises being in a distorted point, since this makes it harder to find improvements. Finding the right trade-off is the heart of our analysis of \dydisOM. For $x\in \{0,1\}^n$ and some suitably chosen constant $\delta>0$ we define}
\inShortVersion{We consider the following potential function for $x\in \{0,1\}^n$ and some suitably chosen constant $\delta>0$:}
\begin{align}
    P(x) := \ZM(x) + \ind{x\notin\CC_s}\frac\delta{\lambda p} \E\big[Y^\ast(\ZM(x))\big].    \label{eq:potential}
\end{align}
We will compute bounds on the drift 
\begin{align}
\Delta(x) := \E\big[P(x^\sss{(t)}) - P(x^\sss{(t+1)}) \mid x^\sss{(t)} = x\big]. \label{eq:drift}
\end{align}
Note that by our sign convention, a positive drift corresponds to progress towards smaller potentials, and thus we want to compute a positive lower bound on $\Delta(x)$ since we aim to establish an upper bound on the running time.
We comment briefly on the second term in the potential $P(x)$, which balances two effects: on the one hand it is sufficiently small so that the event that a distorted offspring is found from a clean point (which happens with probability at most $\lambda p$) yields a small negative contribution to the drift; on the other hand it is sufficiently large so that the drift from a distorted point is of the same order as the drift from a clean point, even though the probability of making a jump is much smaller.

We state four lemmas, then show that an upper bound on $\Tcom$ follows, and eventually \inLongVersion{prove the new lemmas}\inShortVersion{give a sketch of one of the new lemmas}.
The next lemma is due to \citet{GiessenW17}.
\begin{lemma}[{\cite[Lemma 4]{GiessenW17}}]\label{lem:bindrift}
    Let, for some $k\ge 1$, $Y_1,\ldots,Y_\lambda\sim\mathrm{Bin}(k, 1/n)$ be iid binomial random variables, and define $Y^\ast(k):=\max(Y_1,\ldots,Y_\lambda)$. It follows that 
    \begin{enumerate}
        \item if there exists $\alpha>0$ with $\alpha=O(1)$ such that $k=n/(\ln^\alpha\lambda)$, and $\lambda=\omega(1)$, then 
        \begin{equation}
            \E[Y^\ast(k)] = (1\pm o(1))\frac{\ln(\lambda)}{\ln\ln(\lambda) + \ln(n/k)},\nonumber
        \end{equation}
        \item for all $k$, we have $\E[Y^\ast(k)]\ge \lambda k/(\lambda k + n)$,
        \item if $k=o(n/\lambda)$, then $\E[Y^\ast(k)]=(1-o(1))\lambda k/n$.
    \end{enumerate}
\end{lemma}
 The following lemma provides useful bounds for the drift analysis. \inLongVersion{We postpone the proof.} \inShortVersion{We omit the proof.}
\begin{lemma}\label{lem:bindriftdiff}
    Consider the setting of Lemma~\ref{lem:bindrift} under Assumption~\ref{ass:main}. For $k\ge k^\ast$ it holds that 
    \begin{enumerate}
        \item $\E[Y^\ast(k+\ln n)]= (1+o(1))\E[Y^\ast(k)]$;
        \item $\ln(n)\lambda p = o(\E[Y^\ast(k)])$;
        \item $\ln(n)q=o(\E[Y^\ast(k)])$;
        \item $1/n=o(\E[Y^\ast(k)])$;
        \item for all $k\le n/\lambda$, we have $\E[Y^\ast(k)]=\Omega\big(P(x)\tfrac\lambda n\big)$.
    \end{enumerate}
\end{lemma}

The next lemma obtains bounds on the drifts and a non-sharp upper bound on $\Tcom$, i.e., it contains an additional $\ln$-factor compared to Theorem~\ref{thm:main}. Our lower bound on the drift matches the drift of the \ocl and \onemax with the same parameters (using the \zeromax potential for those cases). \inLongVersion{We postpone its proof.}%
\begin{lemma}\label{lemma:dynRun}
Under Assumption \ref{ass:main}, we have $\Delta(x)=\Omega(\E[Y^\ast(k)])$ for all $x$ with $k\ge k^\ast$. Moreover, $\Tcomdynall\le n \ln^2 n$ w.h.p. 
\end{lemma}%
\inShortVersion{We omit the proof. It is based on a careful case analysis (distinguishing whether there is a clone of the parent and whether there is a distorted offspring) to bound the drift and obtain a lower bound on $\E[\Tcomdyn]$, and Markov's inequality to obtain a tail bound.}

The following lemma shows that even though the noise in \dydisOM is dynamic, these dynamics are not seen by the algorithm since each sampled point always returns the same function value w.h.p. Moreover, it shows that for all $t\le \Tcom$, the algorithm never jumps to a search point with much smaller $\OM$-value than the current search point $x^\sss{(t)}$. \begin{lemma}\label{lemma:dynToStat}
Under Assumption \ref{ass:main}, for \dydisOM it holds w.h.p.\ that if there exists $s\!<\! t\!\le\!\Tcomdyn$ such that $y^\sss{(s)}\!=\!y^\sss{(t)}$, then $\dydisOM(y^\sss{(s)})\!=\!\dydisOM(y^\sss{(t)})$. Moreover, for any $\varepsilon>0$ and $n$ sufficiently large, w.h.p.\
\begin{equation}
    \big\{\forall t<t'\le \lceil\Tcom/\lambda\rceil: \OM(x^\sss{(t')}) \ge \OM(x^\sss{(t)})-\varepsilon \ln^4(n)\big\}.\nonumber
\end{equation}
\end{lemma}
We postpone \inShortVersion{a sketch of }%
the proof and first prove the upper bound on $\Tcom$, assuming  \inShortVersion{Lemma~\ref{lemma:dynToStat}}\inLongVersion{Lemmas \ref{lem:bindriftdiff}--\ref{lemma:dynToStat}}.
\begin{proof}[Proof of Theorem~\ref{thm:main}, upper bound on $\Tcom$.]
$ $\newline
We consider a run of the algorithm on \dydisOM, and make the connection to \disOM at the end of the proof. 
\inLongVersion{Since the drift of the potential is similar to the drift of the \ocl and the \opl on \onemax, this part is similar to previous analyses of those situations~\cite{Doerr2013,GiessenW17}. } 
We split the run of the \ocl on \dydisOM into two phases. Let $T_1$ denote the number of function evaluations until the first time that the \ocl moves to a point $X_{T_1}$ that has $\ZM$-value at most $n/\lambda - \varepsilon\ln^4(n)$, where $\varepsilon$ is as in Lemma~\ref{lemma:dynToStat}. Let $T_2$ denote the number of function evaluations after moving to $X_{T_1}$ until moving to a point with $\ZM$-value at most $k^\ast$. Then $\Tcomdyn\le T_1+T_2$, since a distorted point with $\ZM$-value at most $k^\ast+d$ (which has fitness at least $n-k^\ast$) may have been found before $T_1+T_2$. 
We will argue that w.h.p.\ $T_1, T_2=O(n\ln n)$.

We first consider $T_1$, for which we apply a variable drift theorem and Markov's inequality (the presence of the indicator in the potential in~\eqref{eq:potential} prevents direct use of drift theorems with tail bounds, since the decay on the jump size distribution does not decay exponentially). 
By Lemmas~\ref{lem:bindrift} and \ref{lemma:dynRun}, we obtain that 
\[ 
\Delta(x)\ge \begin{dcases}
    \Omega\Big( \frac{\ln \lambda}{2\ln\ln\lambda}\Big),&\text{if }k\ge n/\ln \lambda, \\
    \Omega(1),&\text{if }k\ge n/\lambda - \varepsilon\ln^4 n.
\end{dcases}
\]
By the variable drift theorem as stated in \cite[Theorem 2.3.3]{lengler2020drift}, the expected number of generations $\E[\lceil T_1/\lambda\rceil]$ is of order at most 
\begin{align}
    n\frac{\ln\ln \lambda}{\ln\lambda} + \frac{n}{\ln\lambda}
 = O\Big(n\frac{\ln\ln\lambda}{\ln\lambda}\Big).\label{eq:phase-1}
\end{align}
Since the number of function evaluations is a factor $\lambda$ larger than the number of generations, it follows by Markov's inequality that 
\begin{equation}
\Prob\big(T_1\ge n\ln n\big) \le \frac{O\Big(n\frac{\ln\ln\lambda}{\ln\lambda}\lambda\Big)}{n\ln n}=O\Big(\frac{\ln\ln\ln n}{\ln\ln n}\Big)=o(1),
\end{equation}
using that $\lambda=\Theta(\ln(n))$ by Assumption~\ref{ass:main}.

Recall that at time $T_1$ the current search point $x$  satisfies $\ZM(x)\!\le \!n/\lambda \!- \!\varepsilon\ln^4n$. By Lemma~\ref{lemma:dynToStat}, w.h.p.\ the \ocl never visits a search point $x'$ with $\ZM(x')\!>\!n/\lambda$ after time $T_1$.
By Lemma~\ref{lemma:dynRun} and Lemma~\ref{lem:bindrift}(5), we have for all $x$ with $\ZM(x)\le n/\lambda$ that
\begin{align*}
    \Delta(x) = \Omega\big(\E[Y^\ast(k)]\big)=\Omega(P(x) \cdot \lambda/n),
\end{align*}
which corresponds to a multiplicative drift for all the visited search points after $T_1$ w.h.p. By the multiplicative drift theorem with tail bounds~\cite[Theorem~5]{DoerrG13}, w.h.p.\ the number of generations to reach a point with $\ZM(x)\le k^\ast$ from $X_{T_1}$ is at most $O(\ln(n) \cdot n/\lambda)$. The number of function evaluations in this phase is by a factor $\lambda$ larger, namely $O(n\ln n)$, as required. 
Combined with \eqref{eq:phase-1} this implies that $\Tcomdyn=O(n\ln n)$ w.h.p.

We will now translate this bound on \dydisOM (with dynamic noise) to a  bound on \disOM (with frozen noise).
For \dydisOM and \disOM, each point is distorted at the first time it is sampled with probability $p$ independently of other points. Consequently, $x^\sss{(0)}$ is clean w.h.p. Moreover, by Lemma~\ref{lemma:dynToStat} w.h.p. the function values of  points sampled in \dydisOM never change, so the runs on the dynamic and the `frozen' model are identical. Hence, the upper bound in \eqref{eq:maincom} on $\Tcomall$ follows from $\Tcomdynall=O(n\ln n)$. 
\end{proof}

\inLongVersion{
It is left to prove Lemmas~\ref{lem:bindriftdiff}--\ref{lemma:dynToStat}.
\begin{proof}[Proof of Lemma~\ref{lem:bindriftdiff}]
\emph{Part 1.} The lower bound $\E[Y^\ast(k+\ln(n))]\ge \E[Y^\ast(k)]$ is trivial, since $\E[Y^\ast(k)]$ is increasing in $k$. 
 We first show a helping statement for the upper bound. Let $0\le k_1<k_2$, we show that for two independent random variables $Y^\ast(k_1)$ and $Y^\ast(k_2)$ it holds that $\E[Y^\ast(k_1+k_2)]\le \E[Y^\ast(k_1)+ Y^\ast(k_2)]$.
Let $Z_{i,j}\sim\mathrm{Ber}(1/n)$ be iid Bernoulli random variables for $i, j\ge 1$, so we may couple the random variables as follows
    \begin{align*}
     Y^\ast(k_1+k_2)&=\max_{i\le \lambda}(Z_{i,1}+\dots+Z_{i, k_1+k_2}), \\
     Y^\ast(k_1)&=\max_{i\le \lambda}(Z_{i,1}+\dots+Z_{i, k_1}), \\
     Y^\ast(k_2)&=\max_{i\le \lambda}(Z_{i,k_1+1}+\dots+Z_{i, k_2}).
    \end{align*}
    For any $s$, 
    \begin{align*}
        \Prob\big(Y^\ast(k_1+k_2) &- Y^\ast(k_1)\ge s\big)\\&\le \Prob\big(\exists i\le \lambda: Z_{i, k_1+1} +\dots+Z_{i, k_1+k_2}\ge s\big)\\ &= \Prob\big(Y^\ast(k_2) \ge s\big).
    \end{align*}
    Consequently, $\E[Y^\ast(k+\ln n)]\!\le\!\E[Y^\ast(k)]+ \E[Y^\ast(\ln n)]$ by stochastic domination and linearity of expectation. Part 1 follows if we show that $\E[Y^\ast(\ln n)]\!=\!o\big(\E[Y^\ast(k)]\big)$ for all $k\!\ge\! k^\ast$. This follows by Lemma~\ref{lem:bindrift}(3) for $k_{\ref{lem:bindrift}}\!=\!\ln n$, using that $k^\ast\!=\!n^{\Omega(1)}$ by Assumption~\ref{ass:main}. 
    
    \emph{Part 2-4}. These follow immediately from Lemma~\ref{lem:bindrift}(3) and Assumption~\ref{ass:main}.

    \emph{Part 5.} We substitute $P(x)$ from~\eqref{eq:potential} to observe that it suffices to show that
   \[
   \E[Y^\ast(k)] = \Omega\big((k\lambda/ n)  + \tfrac{1}{np}\E[Y^\ast(k)]\big).
   \]
   Since $p=\omega(1/n)$, the second term on the right-hand side is of smaller order than $\E[Y^\ast(k)]$, and the above equality is satisfied by Lemma~\ref{lem:bindrift}(2) under the assumption that $k\le n/\lambda$.
\end{proof}
}
\inLongVersion{
\begin{proof}[Proof of Lemma \ref{lemma:dynRun}]
We will establish a lower bound on the drift $\Delta(x)$ defined in~\eqref{eq:drift}. Define 
\begin{align*}
\Delta^+(x) &:= \E[\max\{P(x^\sss{(t)}) - P(x^\sss{(t+1)}), 0\}\mid x^\sss{(t)} = x], \\ 
\Delta^-(x) &:= \E[\max\{P(x^\sss{(t+1)}) - P(x^\sss{(t)}), 0\}\mid x^\sss{(t)} = x],
\end{align*}
so that 
\begin{align}
    \Delta(x) = \Delta^+(x) - \Delta^-(x).
\end{align}
We will first obtain lower bounds on $\Delta^+(x)$,  then obtain upper bounds on $\Delta^-(x)$, distinguishing in both cases between clean and distorted points $x$. Then we combine the bounds and find $\Delta(x)\!=\!\Omega(\E[Y^\ast(k)])$ under Assumption~\ref{ass:main}. Eventually we obtain a non-matching tail bound on $\Tcom$.

\emph{Forward progress, clean points.} First, assume that $x$ is a clean point with $\ZM(x)=k$. Let $X_i$ denote the number of bits flipped from $1$ to $0$ for the $i$-th offspring (with $i\le \lambda$), similarly let $Y_i$ be the number of bits flipped from $0$ to $1$ for the $i$-th offspring. Moreover, we write $Y^\ast(k):=\max_i(Y_i)$ and $i^\ast$ for the index such that $Y_{i^\ast}=Y^\ast(k)$ (with arbitrary tie-breaking rule). Let $\CE_1$ be the event that all offspring are clean.
Then, 
\begin{align}
    \Delta^+(x) \ge
    \E[Y^\ast(k)\ind{X_{i^\ast}=0}\ind{\CE_1}],
\end{align}
considering only the case where all offspring are clean, and in which the $i^\ast$th offspring has no 1-bits flipped into a 0-bit. 
Applying the law of conditional probability twice,  using that $Y^\ast(k)$ is determined by all flipped $0$-bits in the offspring, and $X_{i^\ast}$ is determined from all mutations,  
we obtain
\begin{equation}\nonumber
    \Delta^+(x)\!
    \ge\!
    \E\Big[Y^\ast(k)\E\big[\ind{X_{i^\ast}\!=\!0}\Prob\big(\CE_1\!\mid\! \mbox{flipped bits}\big)\!\mid\! \mbox{flipped 0-bits}\big]\Big].
\end{equation}
Since already visited clean points remain clean, and newly visited points are distorted with probability $p$, by a union bound $\Prob\big(\CE_1\!\mid\! \mbox{flipped bits}\big)\ge 1-\lambda p=(1-o(1))$. By independence, the probability that no $1$ bits are flipped is at least $(1-1/n)^{n-k}\ge 1/e$. So we obtain for clean $x$ that 
\begin{align}
    \Delta^+(x)
    &\ge
    (1-o(1))\E[Y^\ast(k)]/e.\label{eq:positive-drift-clean}
\end{align}

\emph{Forward progress, distorted points.} For distorted points we consider the additional event $\CE_2$ that there is no clone among the offspring, and the event $\CE_3$ that there exists $i$ such that $Y_i\le 1$. As a result, on the event $\CE_1\cap\CE_2\cap\CE_3$ there is no clone of the parent, all offspring are clean, and $\ZM(x^\sss{(t+1)})-\ZM(x^\sss{(t)})\le 1$. Hence, 
\begin{align*}
\Delta^+(x)
&\ge 
\E[\big(P(x^{(t)}) - P(x^{(t+1)})\big)\ind{\CE_1}\ind{\CE_2}\ind{\CE_3}\mid x^{(t)} = x] \\
&\ge \big(\delta/(\lambda p)\cdot\E[Y(k)^\ast]-1\big)\cdot\Prob\big(\CE_1\cap\CE_2\cap\CE_3\big)\\
&\ge \delta/(2\lambda p)\cdot\E[Y(k)^\ast]\cdot\Prob\big(\CE_1\cap\CE_2\cap\CE_3\big),
\end{align*}
since $\E[Y_k^\ast]/(\lambda p)=\omega(1)$ by Lemma~\ref{lem:bindriftdiff}(2).
We focus on $\Prob\big(\CE_1\cap\CE_2\cap\CE_3\big)$. Since $\ind{\CE_1\cap\CE_3}$ is determined by all the bit flips in the offspring, we obtain by the law of total probability and a union bound over the possibly distorted offspring that
\begin{align*}
\Prob\big(\CE_1\cap\CE_2\cap\CE_3\big) &= \E\big[\ind{\CE_1\cap\CE_3}\cdot\Prob\big(\CE_2\mid \mbox{flipped bits}\big)\big] \\
&\ge (1-\lambda p)\cdot\Prob\big(\CE_1\cap\CE_3\big) \\&\ge (1-o(1))\cdot\big(\Prob(\mbox{no clone}) - \Prob(\forall i\le\lambda: X_i\ge2)\big),
\end{align*}
applying a union bound and substituting the definition of $\CE_1$ and the complement of $\CE_3$ in the second inequality. By Lemma~\ref{lem:probability-of-no-clone} and using that all $X_i+Y_i\sim\mathrm{Bin}(n,1/n)$ are iid, it follows that for $n$ sufficiently large (using $\lambda=\omega(1)$ and $1/\eta=0.63..>1/2$)
\begin{align}
    \Prob\big(\CE_1\cap\CE_2\cap\CE_3\big)
    &\ge 
    (1-o(1))\bigg((1+o(1))(1/\eta)^{\lambda} - \left(\binom{n}{2}\frac{1}{n^2}\right)^\lambda\bigg) \nonumber\\
    &\ge (1-o(1))\big((1+o(1))(1/\eta)^\lambda - \left(1/2\right)^\lambda\big) \nonumber\\
    &\ge q/2,\nonumber
\end{align}
since $q=\eta^{-\lambda}$ by definition.
We obtain for $n$ large that
\begin{align}
    \Delta^+(x) \ge \delta\E[Y^\ast_k]\frac{q}{4\lambda p}\ge \E[Y^\ast(k)],\label{eq:forward-distorted}
\end{align}
since $p\lambda=O(p\ln(n))=o(q)$ by Assumption~\ref{ass:main}. 

Combined with~\eqref{eq:positive-drift-clean}, this implies 
$\Delta^+(x) \ge \E[Y^\ast(k)]/(2e)$ for both clean and distorted $x$ when $n$ is sufficiently large.

We will now establish upper bounds on $\Delta^-(x)$.

\emph{Backwards progress, clean points.}
Let $\CE_4$ be the event that there exists a distorted offspring, $\CE_5$ be the event that all offspring flip at most $\ln(n)$ bits, and recall the event $\CE_2$ is that there is no clone of the parent.
We distinguish the cases whether $\CE_2$, $\CE_4$, and $\CE_5$ hold. Abbreviating $r^+:=\max\{r, 0\}$ and $P_t=P(x^\sss{(t)})$, we obtain 
\begin{equation} 
\begin{aligned}
\Delta^-(x)&= \E\big[(P_{t+1}-P_t)^+\ind{\CE_4}\ind{\CE_5}\big]\\
&\hspace{15pt}+\E\big[(P_{t+1}-P_t)^+\ind{\neg\CE_2}\ind{\neg\CE_4}\ind{\CE_5}\big]\\
&\hspace{15pt}+\E\big[(P_{t+1}-P_t)^+\ind{\CE_2}\ind{\neg\CE_4}\ind{\CE_5}\big]\\
&\hspace{15pt}+\E\big[(P_{t+1}-P_t)^+\ind{\neg\CE_5}\big].
\end{aligned}\label{eq:neg-drift-4-cases}
\end{equation}
For the first term, the indicator in \eqref{eq:potential} is $0$ for $x^\sss{(t)}$ and is potentially $1$ for $x^\sss{(t+1)}$, in which case the additional term is at most $(\delta/(\lambda p))\cdot\E[Y^\ast(k+\ln(n))]$, using that the expectations $\E[Y^\ast(k)]$ are increasing in $k$, and that on $\CE_5$ the maximal jump size is bounded from above by $\ln(n)$. Since $\E[Y^\ast(k+\ln(n))]=(1+o(1))\E[Y^\ast(k)]$ by Lemma~\ref{lem:bindriftdiff},
\begin{align}
    \E\big[(P_{t+1}&-P_t)^+\ind{\CE_4}\ind{\CE_5}\big] \nonumber\\&\le \big((\ln(n) +  (\delta/(\lambda p))\cdot\E[Y^\ast(k+\ln(n))]\big)\cdot\Prob\big(\CE_4) \nonumber\\
    &\le \ln(n)\lambda p+ 2\delta \cdot \E[Y^\ast(k)],\label{eq:one-clean}
\end{align}
using also $\Prob\big(\CE_4\big)\!=\!\Prob\big(\exists\mbox{distorted offspring}\big)\!\le\! \lambda p$ by a union bound.

The second term in \eqref{eq:neg-drift-4-cases} equals $0$: there is no distorted offspring on $\neg\CE_4$, and a clone of the parent on $\neg\CE_2$, so the $\ZM$ value does not increase and the indicators in~\eqref{eq:potential} are both $0$. 

For the third term in \eqref{eq:neg-drift-4-cases} we observe that the potential can increase by at most $\ln(n)$ since there is no distorted offspring. On the event $\CE_2$ there is no clone of the parent, so by Lemma~\ref{lem:probability-of-no-clone}
\begin{align}
    \E\big[(P_{t+1}&-P_t)^+\ind{\CE_2}\ind{\neg\CE_4}\ind{\CE_5}\big] \nonumber\\
&\le    \ln(n)\cdot\Prob\big(\nexists \mbox{clone}\big) = (1+o(1))\ln(n)q.\label{eq:three-clean}
\end{align}
For the fourth term in~\eqref{eq:neg-drift-4-cases} we use that the maximal difference in potential between two search points is $n+\delta\E[Y^\ast_n]/(\lambda p)=O(n^2)$ by Lemma~\ref{lem:bindrift}(1), using also that $p=\omega(1/(n\ln n))$ by Assumption~\ref{ass:main}. 
By Lemma~\ref{lem:hamming-log} and a union bound over the $\lambda$ offspring
\begin{align}
    \E\big[(P_{t+1}&-P_t)^+\ind{\neg\CE_5}\big] \label{eq:four-clean} \\&\le O(n)\cdot \lambda\cdot \Prob\big(\mbox{offspring flips $\ge \ln(n)$ bits}\big)=o(1/n).\nonumber
\end{align}

Substituting the bounds \eqref{eq:one-clean}--\eqref{eq:four-clean} into \eqref{eq:neg-drift-4-cases} and recalling that the second term in \eqref{eq:neg-drift-4-cases} is 0, we obtain for clean $x$ that 
\begin{equation}
    \Delta^-(x)\le 2\delta \E[Y^\ast(k)]+ O(\ln(n)\lambda p) + O(\ln(n)q) +o(1/n).
\end{equation}
Combining this bound with the forward drift $\Delta^+(x)$ from~\eqref{eq:positive-drift-clean}, and substituting the bounds from Lemma~\ref{lem:bindriftdiff}(2--4), it follows that for $\delta>0$ sufficiently small
\begin{equation}
    \Delta(x) \ge \Omega\big(\E[Y^\ast(k)]\big).
\end{equation}

\emph{Backwards progress, distorted points.}
We first consider the case when there is a clone. If the selected offspring is again distorted, then it must have the same or a smaller $\ZM$-value than $x$ to be accepted, and the indicator in~\eqref{eq:potential} does not increase since the expectations $\E[Y^\ast(k)]$ are increasing in $k$. If the selected offspring is \emph{not} distorted, then it must also have smaller $\ZM$-value, and the indicator decreases. Overall, the potential cannot increase when there is a clone of $x$. Hence, 
\begin{equation}
    \Delta^-(x)=\E[(P_{t+1}-P_t)^+\ind{\CE_2}].
\end{equation}
Distinguishing whether there is an offspring that flips more than $\ln(n)$ bits (event $\CE_5$), we obtain by the same reasoning as in \eqref{eq:four-clean}
\begin{align}
    \Delta^-(x)&\le \E[(P_{t+1}-P_t)^+\ind{\CE_2}\ind{\CE_5}] + \E[(P_{t+1}-P_t)^+\ind{\neg\CE_5}] \nonumber\\
    &\le \E[(P_{t+1}-P_t)^+\ind{\CE_2}\ind{\CE_5}] + o(1/n).\label{eq:neg-dist}
\end{align}
On the event that there are at most $\ln(n)$ bit flips, considering the worst case that the selected offspring is distorted, we obtain by Lemma~\ref{lem:bindriftdiff}(1)
\begin{align}
P_{t+1}-P_t &\le \ZM(x^\sss{(t+1)})-\ZM(x^\sss{(t)}) \nonumber\\
&\hspace{15pt}+ (\delta/(\lambda p))\cdot\E[Y^\ast(\ZM(x^\sss{(t+1)}))-Y^\ast(\ZM(x^\sss{(t)}))] \nonumber\\
&\le \ln(n) + (\delta/(\lambda p))\cdot o\big(\E[Y^\ast(\ZM(x^\sss{(t)}))]\big)
.\nonumber
\end{align}
Substituting this back into~\eqref{eq:neg-dist} and using that  by Lemma~\ref{lem:probability-of-no-clone} $\Prob\big(\CE_2\big)=\Prob\big(\mbox{no clone}\big)\le 2q$, we obtain by Lemma~\ref{lem:bindriftdiff}(1)
\begin{align}
    \Delta^-(x)&\le 2q\cdot \big(\ln(n) + (\delta/(\lambda p))\cdot o\big(\E[Y^\ast(k)]\big)\big) + o(1/n).\nonumber
\end{align}
By Lemma~\ref{lem:bindriftdiff}, we have $q\ln(n), 1/n=o\big(\E[Y^\ast(k)]\big)$. Recalling the lower bound on $\Delta^+(x)$ from~\eqref{eq:forward-distorted}, we obtain
\begin{align*}
    \Delta(x)&\ge \frac{q\delta}{\lambda p}\Big(\tfrac14\E[Y^\ast_k]-o\big(\E[Y^\ast(k)]\big)\Big) - o\big(\E[Y^\ast(k)]\big) \\
    &=\Omega\big(\tfrac{q}{\lambda p}\E[Y^\ast_k]\big)=\Omega(\E[Y^\ast_k]),
\end{align*}
using that $q/(\lambda p)=\omega(1)$ by Assumption~\ref{ass:main},
finishing the proof of the first part of Lemma~\ref{lemma:dynRun}. 

\emph{Upper bound $\Tcomdyn$}. We will now obtain an upper bound on $\Tcomdyn$ that holds w.h.p. By the first part of the lemma and Lemma~\ref{lem:bindrift} we have $\Delta(x)=\Omega(\min\{1,\lambda k/n\})$. By the variable drift theorem as formulated in \cite[Theorem 2.3.3]{lengler2020drift}, we obtain for the number of generations $\lceil\Tcomdyn/\lambda\rceil$ that $\E\big[\lceil\Tcomdyn/\lambda\rceil\big]=O(n)$ for $\lambda=\Theta(\ln(n))$. 
The number of function evaluations is a factor $\lambda$ larger.
Hence, by Markov's inequality
\[
\Prob\big(\Tcom\ge n\ln^2(n)\big)\le \frac{O(n\ln(n))}{n\ln^2(n)}=o(1).\qedhere
\]
\end{proof}
}%

\inShortVersion{We omit the proofs of Lemmas~\ref{lem:bindriftdiff}--\ref{lemma:dynRun}, but provide a sketch of the proof of Lemma~\ref{lemma:dynToStat}.}
\begin{proof}[Proof\inShortVersion{ sketch} of Lemma \ref{lemma:dynToStat}]
\inLongVersion{We first verify the first part of Lemma~\ref{lemma:dynToStat}, and verify
the second statement at the end.}
\inShortVersion{We verify the first statement of Lemma~\ref{lemma:dynToStat}, and omit the proof of the second statement.}
The only way there could exist $t'\!<\!t$ such that $y^\sss{(t')}\!=\!y^\sss{(t)}$, but $\dydisOM(y^\sss{(t')})\!\neq\!\dydisOM(y^\sss{(t)})$, is when $y^\sss{(t')}$ is distorted and resampled at a later time $t$ at which it is clean (clean points remain clean). Hence, it suffices to argue that w.h.p.\ the algorithm never resamples a point that was distorted at the first time it was sampled. To do so, we will analyse the positive and negative jumps of the \ocl on \dydisOM in polynomial time intervals. 

Let $s:=C(n/k^\ast)\ln^4(n)$ for some large constant $C>0$, and define for the $t$-th function evaluation the random times $t_1(t),\ldots,t_s(t)$, which are the first $s$ unique evaluations after the $(t-1)$-st evaluation from a clean parent (for convenience we assume $s\in\N$). \inLongVersion{We will show }%
\inShortVersion{It can be shown }%
that there exists $C', \varepsilon>0$ such that w.h.p.\ the following three events hold: 
\begin{itemize}
    \item[(i)] $\forall t\le n^2$, the total time in the interval $[t, t_s(t)]$ at which the parent is distorted is at most $O(\ln^3(n)/q)$.
    \item[(ii)] $\forall t\le n^2, r\in[t_s(t), \Tcomdyn]$, each sampled point $y^\sss{(r)}$ satisfies $\OM(y^\sss{(r)})> \OM(x^\sss{(\lceil t/\lambda\rceil-1 )})+C'\ln(n)$.    
    \item[(iii)] the number of sampled distorted points until the fixed target is reached is $O(pn\ln(n))$.
\end{itemize}
We \inLongVersion{first }%
show \inShortVersion{here }%
that the statement follows under the assumption that these events hold w.h.p. 
Recall that it is sufficient to bound the probability that a distorted point is resampled. If an offspring $y^\sss{(t)}$ with $\OM(y^\sss{(t)})=k$ is distorted, then by event (i) its parent (which is $x^\sss{(\lceil t/\lambda\rceil -1 )}$) has $\OM$-value at least $k-C'\ln(n)$. By event (iii), the only times at which the distorted point $y^\sss{(t)}$ can be resampled is during the interval $[t, t_s(t)]$.
At each time $r\in[t, t_s(t)]$ the probability that $y^\sss{(t)}$ is resampled is at most $1/n$. By a union bound over the precisely $s=\Theta((n/k^\ast)\ln^4(n) )$ times in $[t, t_s(t)]$ at which the parent is clean (definition of $t_1(t),\ldots, t_s(t)$), and the $O(\ln^3(n)/q)$ times at which the parent is distorted (event (i)), the probability that the point $y^\sss{(r)}$ is resampled in this interval is at most 
\[
O(\max\{\ln^4(n)/k^\ast, \ln^3(n)/(nq)\}).
\]
By a union bound over the at most $O(pn\ln(n))$ distorted points visited (event (iii)), no distorted point is resampled with probability at least 
\[
\begin{aligned}
1-O&(pn\ln(n)\cdot \max\{\ln^4(n)/k^\ast, \ln^3(n)/(nq)\})  \\
&=1-O(\max\{\ln^5(n)\cdot p\cdot n/k^\ast, \ln^4(n)p/q\}) = 1-o(1), 
\end{aligned}
\]
since $p\le k^\ast/n^{1+\delta} = o(k^\ast /(n\ln^5(n)))$ and $p\le qn^{-\delta}=o(q/\ln^4(n))$ by Assumption \ref{ass:main}, see~\eqref{eq:ass-rewritten}. 
Thus, the first lemma statement follows if \inLongVersion{we show that }%
the three events hold with high probability. \inShortVersion{\qedhere}

\inLongVersion{
\textbf{Preparations events (i-ii)}.
We will analyse the \ocl on \dydisOM during intervals $[t, t_s(t)]$ to obtain bounds for events (i-ii).  
We start with the number of OM-improving steps in $[t, t_s(t)]$. Following the proof of Lemma \ref{lemma:dynRun}, the probability of an OM-improving step (for $r\in[t, t_s(t)]$ such that $r\le\Tcomdyn$) is at least $c\lambda k^\ast /n$ for some small constant $c>0$, independently of the history. So the expected number of OM-improving steps is at least $(s/\lambda)\cdot c \lambda k^\ast /n =\Omega(s\cdot k^\ast /n)=\Omega(\ln^4(n))$, considering only OM-improving steps from clean points. By a Chernoff bound it follows that w.h.p.\ for all $t\le n^2$ the number of OM-improving generations in $[t, t_s(t)]$ is at least $C'\ln^4(n)$ for some constant $C'$ that depends on the sufficiently large constant $C$ in the definition of $s$.

We move on to the `progress' away from the target during the interval $[t, t_s(t)]$. 
We will bound the total number of times that the algorithm jumps backwards from above, distinguishing between jumps from/to either clean or distorted points.  

\emph{Clean to clean.} We start with backward jumps from clean points to other clean points, whose total size is bounded from above by $O(\ln(n))$ times the number of generations that there is no clone of the parent (the upper bound on the jump size comes from Lemma~\ref{lem:hamming-log}). Since there is no clone of the parent with probability at most $2q$ by Lemma~\ref{lem:probability-of-no-clone}, the number of generations without clone is stochastically dominated by a Binomial random variable $\mathrm{Bin}(s/\lambda, 2q)\preccurlyeq\mathrm{Bin}(\max\{s/\lambda, \ln(n)/(2q)\}, 2q)$. 

By Chernoff bounds, the probability that for a fixed $t$ the number of non-clone generations in $[t, t_s(t)]$ exceeds $5\max\{sq/\lambda, \ln(n)\}$ is at most 
\[
\exp\big(-4\max\{s/\lambda, \ln/q\}\cdot q\big)=n^{-4},
\]
using that $q=O(k^\ast/(n\ln^2(n)))$ and $\lambda=\Theta(\ln(n))$ by Assumption~\ref{ass:main}, and $s=\Theta(\ln^4(n)n/k^\ast)$ by definition. 
By a union bound, w.h.p.\ there is no $t\le n^2$ such that at the times $t_1(t)$,$\ldots$, $t_s(t)$ the number of non-clone generations exceeds $5\ln(n)$.
The jump sizes away from the optimum from all such jumps sum up to at most $O(\ln^2(n))$.

\emph{Clean to distorted, distorted to clean}. The number of distorted offspring sampled at the times $t_1(t)$,$\ldots$, $t_s(t)$ (at which the parent is clean by assumption) is stochastically dominated by $\mathrm{Bin}(s, p)\preccurlyeq \mathrm{Bin}(\max\{s, \ln(n)/p\}, p\})$, since in \dydisOM previously sampled clean points remain clean, while other points are distorted independently with probability $p$. By Chernoff bounds, w.h.p.\ there is no $t\le n^2$ such that the number of distorted offspring sampled at times $\{t_1(t),\ldots, t_s(t)\}$ exceeds $5\max\{sp, \ln(n)\}=O(\ln(n))$, since $p=o(k^\ast/(n\ln^3(n))$ by Assumption \ref{ass:main}. 
When jumping from a clean point to a distorted point, by  Lemma~\ref{lem:hamming-log}, the $\OM$-value decreases by at most $O(\ln(n))$. Hence, the total decrease of OM-value during the interval $[t, t_s(t)]$ is $O(\ln^2(n))$.
The number of jumps from a distorted point to a clean point is also $O(\ln(n))$, since it is bounded from above by the number of sampled distorted points from a clean parent, and at each step the decrease is at most $O(\ln(n))$.

\emph{Distorted to distorted}. To analyse the number of jumps between distorted points, we establish an upper bound on the total number of jumps between distorted points before jumping to a clean point. The probability of moving between two generations is bounded from below by $q/2$, considering only the case in which there is no clone of the parent. The probability of moving from a distorted point to a distorted point is at most $\lambda p$. Hence, the probability of jumping from a distorted point to another distorted point at its next move is bounded from above by $2\lambda p/q\le 1/2$ (using that  $p\ln(n)/q=o(1)$ by Assumption \ref{ass:main}), independently of the past. Thus, the number of consecutive moves between distorted points is stochastically dominated by a geometric random variable $\mathrm{Geo}(1/2)$. Hence, the probability that there are $6\ln(n)$ consecutive moves between distorted points starting from a distorted point is at most 
\begin{equation}
(1-1/2)^{3\ln(n)/(1/2)}\le \exp(-3\ln(n)). \label{eq:concentration-geo}
\end{equation}
By a union bound over the at most $n^2$ distorted points, this event does not occur for any of the first $n^2$ distorted points w.h.p. 
Combined with the at most $O(\ln(n))$ sampled distorted offspring from clean points, the total number of visited distorted points in the time interval $[t, t_s(t)]$ is at most $O(\ln^2(n))$, and at each jump between distorted points away from the optimum is at most $O(\ln(n))$ by Lemma~\ref{lem:hamming-log}.

\textbf{Event (i)}. 
As noted in the preparatory reasoning, w.h.p.\ the number of visited distorted points during $[t, t_s(t)]$ is $O(\ln^2(n))$. The number of generations to leave a distorted point is stochastically dominated by a $\mathrm{Geo}(q/2)$ random variable, where $q/2$ is a lower bound on the probability that there is no clone of the parent in one generation. Similar to \eqref{eq:concentration-geo}, w.h.p.\ none of the first $n^2$ times that we visit a distorted point, the algorithm stays longer in the distorted point than $O(\ln(n)/q)$ generations. Since the total number of visited distorted points is $O(\ln^2(n))$ w.h.p., and $\lambda=\Theta(\ln(n))$, event (i) holds w.h.p.

\textbf{Event (ii)}. The total negative progress from clean parents is at most $O(\ln^2(n))$ by the preparatory reasoning, and the total negative progress from distorted parents is $O(\ln^3(n))$ by Lemma~\ref{lem:hamming-log}.
Since the number of OM-improving steps in each interval $[t, t_s(t)]$ is $\Omega(\ln^4(n))$, it follows that there exists $\varepsilon>0$ such that w.h.p.\
\begin{itemize}
\item[(iiia)] $\forall t\le n^2$, the progress in time $[t, t_s(t)]$ is at least $2\varepsilon \ln^4(n)$.
\end{itemize}
Let $t'=t_s(t)$ for some $t$ such that $y^\sss{(t)}$ is distorted.
By the preparatory reasoning, during the interval $[t', t_s(t')]$, the total number of backward jumps that have size at most $O(\ln(n))$ 
is bounded by $O(\ln^2(n))$ w.h.p.
By event (i), w.h.p.\ for any $\varepsilon>0$, each offspring $y^\sss{(r)}$ sampled during the time interval $[t', t_s(t')]$ satisfies 
\begin{align}
\OM(y^\sss{(r)})&\ge \OM(x^{(\lceil t'/\lambda\rceil-1)}) - O(\ln^3(n)) - O(\ln(n))\nonumber\\&\ge \OM(x^{(\lceil t'/\lambda\rceil - 1)}) - \varepsilon \ln^4(n) \nonumber\\ &\!\!\!\overset{(\text{iiia})}\ge \OM(x^{(\lceil t/\lambda\rceil - 1)}) + \varepsilon\ln^4(n).\nonumber
\end{align}
By Lemma~\ref{lem:hamming-log} it follows that $\OM(y^{(t)}) > \OM(x^{(\lceil t/\lambda\rceil - 1)}) - O(\ln(n))$, so $\OM(y^\sss{(r)})\!>\!\OM(y^\sss{(t)})$ for all $r\!\in\![t_s(t), t_s(t_s(t))]$. Iterating this argument for the next intervals yields that event (ii) holds w.h.p.

\textbf{Event (iii)}.
The first time that each sampled point is sampled, it is distorted with probability $p$, independently of the rest. Since a clean point remains clean and $\Tcomdyn=O(n\ln(n))$ w.h.p.\ by Lemma~\ref{lemma:dynRun}, 
 the total number of distorted points is w.h.p.\ stochastically dominated by a binomial random variable $\mathrm{Bin}(C'\ln(n)n, p)$ for some $C'>0$. The first statement of the lemma follows by Chernoff's bound and the reasoning below the definition of the three events. 

The second part follows by reasoning analogous to the reasoning for the reasoning in the proof that event (ii) holds w.h.p.}
\end{proof}

\section{Traps slow down plus strategy}\label{sec:plus-lower}
\begin{proof}[Proof \inShortVersion{Sketch }of Theorem~\ref{thm:main}, lower bound on $\Tplus$]
We consider distances from $\vec 1$ in the two intervals $I:= (2k^\ast,2k^\ast n^\delta]$ and $I' := [k^\ast n^\delta,k^\ast n^{2\delta}]\subset I$ for some sufficiently small $\delta>0$ so that $k^\ast n^{2\delta}\lambda=o(n)$ (which exists by Assumption~\ref{ass:main}). Note that all points in $I$ have fitness less than $n-2k^\ast+d \le n-k^\ast$, so it is necessary to traverse $I$. 
We \inLongVersion{will show}\inShortVersion{claim} that w.h.p., with $\CD$ the set of distorted points,
\begin{enumerate}[(i)]
    \item we visit a distorted point with $\ZM$-value in $I'$,
    \item afterwards the algorithm does not leave $\CD$ for time $n\ln(n)/p$, and 
    \item it takes time $\Omega(n\ln(n)/p)$ to traverse $I$ \inShortVersion{(that has size $|I|=n^{\Omega(1)}$) }%
    within $\CD$ when starting in~$I'$. 
\end{enumerate}
The three items imply the lower bound~\eqref{eq:mainplus}. \inShortVersion{We omit the proofs.}
\inLongVersion{
We prove the items one by one, starting with (i). 
Observe that w.h.p.\ the algorithm does not jump over the interval $I'$, since in the first $O(\ln(n)n/p)=O(\ln(n)n^2)$ number of rounds, each offspring is within Hamming distance $O(\ln(n))$ from its parent by Lemma~\ref{lem:hamming-log}), and $|I'|=\omega(\ln(n))$.
Assume that the algorithm enters $I'$ at a clean point $x^\sss{(0)}$, since otherwise there is nothing to show. 
As long as it does not move to a distorted point, it mimics perfectly the behaviour on \onemax. 
In particular, by Theorem~\ref{the:big-ass-fixed-target-theorem}(1), w.h.p.\ the \opl needs to produce $\Omega(n\ln n)$ offspring to cross the interval $I'$ on \onemax. 
Let us investigate such a run on \onemax further, starting in $x^\sss{(0)}$. Since each mutation has probability $\Omega(1)$ to produce a (Hamming) neighbour of the parent, w.h.p.\ $\Theta(n\ln n)$ offspring are neighbours of their respective parents. 

We will argue next that the set $S$ of those offspring also contains $\Theta(n\ln n)$ \emph{different} individuals. Assume that at some point of the run, the set $S$ has still size $|S|\le n\ln(n)/2$. Then the current parent $x$ has at least $n\ln(n)/2$ neighbours that are not in $S$. Hence, each offspring has probability $\Omega(1)$ to be a neighbour not in $S$. By the Chernoff bound, w.h.p.\ this process produces $\Theta(n\ln n)$ new search points in $\Theta(n \ln n)$ rounds. So w.h.p.\ $|S| = \Theta(n\ln n)$.

Finally, due to the length of the interval, at most $O(k^\ast n^{2\delta})$ generations improve the fitness, so the number of offspring in those generations is at most $O(\lambda k^\ast n^{2\delta}) = o(n\ln n)$. We remove those offspring from $S$, yielding a set $S'$ of size $\Theta(n\ln n)$. 

If we switch to \disOM (but still define $S'$ via the run on \OM), the expected number of \emph{distorted} points in $S'$ is $\Omega(pn) = \omega(1)$. Hence, w.h.p.\ at least one point in $S'$ is distorted.

Recall that the runs on \disOM and \OM are identical up to the point when the first distorted search point is accepted. So unless the \opl accepts some distorted point, it queries the same points in $S'$. However, let $y\in S'$ be distorted with parent $x$. The fitness of $y$ is $\disOM(y) = \OM(y)+d \ge \OM(x)-1+d > \OM(x)$. Since no other search point in the \onemax run in that generation has a higher fitness than $x$, the offspring $y$ will be accepted (or another distorted offspring in the same generation). This proves (i).

For (ii), assume that the algorithm is in a distorted point $x\in I$, and consider a clean offspring $y$ of $x$. The algorithm can only prefer $y$ over $x$ if $\OM(y) = f(y) \ge f(x) = \OM(x)+d$. Hence, in order to accept a clean offspring, we need to decrease the Hamming distance to the origin by at least $d$. Let us for a moment assume that $d$ is constant. Then the probability of decreasing the Hamming distance by at least $d$ from a parent with at most $k^\ast n^{2\delta}$ zeros is at most $O((k^\ast n^{2\delta}/n)^d) = O(n^{2\delta d}(k^\ast/ n)^d)$ where the first term follows from~\cite[Lemma~3]{Paixao2016} and the second one uses that $d = O(1)$. 
By a union bound, the probability that this happens in time $\ln(n)n/p$ is $O(n\ln(n)/p \cdot n^{2\delta d}(k^\ast /n)^d)$. By the bound on $d$ in Assumption~\ref{ass:main} there is $\eps' > 0$ such that
\begin{align*}
(k^\ast/n)^d = e^{-d\ln(n/k^\ast)} \le e^{-(1+\eps')\ln(n/p)} = (p/n)^{1+\eps'}.
\end{align*}
Therefore, the probability that at least one of $\ln(n)n/p$ offspring decreases the Hamming distance by at least $d$ is at most 
\[
O\big((n\ln(n)/p) \cdot n^{2\delta d}(p/n)^{1+\eps'}\big)=O\big(\ln(n)n^{2\delta d}(p/n)^{\varepsilon'}\big)=o(1)
\]
for $\delta=\delta(\varepsilon')>0$ sufficiently small. In other words, w.h.p.\ no clean offspring is accepted for time $\ln(n)n/p$. Obviously, if $d = \omega(1)$ then it is even harder to decrease the Hamming distance by $d$, so the conclusion also holds without the assumption that $d$ is constant.

It remains to show (iii). Consider a modified \opl, which starts in a distorted point in $I'$, but which automatically discards all clean points, regardless of their fitness. Then, to make an improving step within $\CD$, the algorithm needs to query a search point that \emph{(a)} improves the Hamming distance from $\vec 1$ and \emph{(b)} that is distorted. We will show that we can couple the performance with a run on \onemax in which each offspring is discarded with probability at least $1-p$ before considering it for selection. Let us call this a $(1-p)$-rejection run. Indeed, assume the modified \opl runs on \disOM and its current search point is $x \in I$. Assume further that it samples an offspring $y$ which is by an additive term $r > 0$ closer to the optimum. There are two cases. Either $y$ has not been sampled before. In this case, $y$ is clean with probability $1-p$, and thus rejected with this probability. Or $y$ has been sampled before. In this case, $y$ was clean (otherwise we would have moved there earlier), and thus it is rejected with probability $1$. Therefore, for any $r>0$, the probability of moving $r$ closer to the optimum within $\CD$ is at most the probability of moving $r$ closer to the optimum in a $(1-p)$-rejection run on \onemax. Since even the \oea on \onemax takes time $\Omega(n\ln(n))$ to cross $I$ from any point in $I'$ by Theorem~\ref{the:big-ass-fixed-target-theorem}(1), a $(1-p)$-rejection run of the \oea on \onemax  takes time $\Omega(n\ln(n)/p)$. (We need to wait expected time $1/p$ before considering an offspring.) By Theorem~\ref{thm:domination}, the same is true for the \opl. This proves (iii) and concludes the proof of the lower bound on $\Tplus$ in Theorem~\ref{thm:main}.
}
 \end{proof}

\section{Plus strategy still escapes}\label{sec:plus-upper}
The main goal of this section is to \inLongVersion{prove}%
\inShortVersion{sketch the proof of}
Theorem~\ref{thm:largest-factor}. We remark that the results obtained here immediately imply the upper bound on $\Tplus$ in Theorem \ref{thm:main}.
We split the proof of Theorem~\ref{thm:largest-factor} into two cases. 
We start with a rather simple lemma for the case  which covers the first two settings in Theorem~\ref{thm:largest-factor}. 
\begin{lemma}\label{lem:small-p}
    Consider the setting of Theorem~\ref{thm:largest-factor}. There exists a constant $C'>0$ such that w.h.p.\
    \begin{equation*}
    \Tplus = 
    \begin{dcases}
        T,&\text{if }p\cdot T=o(1), \text{ or }\lambda\ge C'\ln(n),\\
        O\big(n\cdot(\lambda + \ln(n/k^\ast))\big),&\text{if }p\cdot n \cdot(\lambda + \ln(n/k^\ast))=o(1).
    \end{dcases}
    \end{equation*}
 \end{lemma}
 \inShortVersion{
 For small values of $p$ we argue that the \opl never encounters a distorted point w.h.p., while for $\lambda\ge C'\ln(n)$ the \ocl mimics the \opl. We omit the details.
 }
 \inLongVersion{
\begin{proof}
For the first case with $pT=o(1)$,
    consider a run of the \opl on \OM for time $T$.
    Then w.h.p.\ fitness target $n-k^\ast$ is reached before $T$ by Theorem~\ref{thm:domination}(a).
    We can couple this run with a run on \disOM until the first time that a distorted search point is queried. By a union bound, the probability that this happens before round $T$ is $p\cdot T= o(1)$. Hence, with high probability the two runs on \OM and on \disOM are identical until time $T$, and w.h.p.\ fitness target $k^\ast$ is reached before then. Hence, w.h.p.\ $\Tplus \le T$.

    For the second case, since we assume that $T\le n^C$, there exists $C'>0$ such that for $\lambda\ge C'\ln(n)$ in the first $n^C$ rounds there is a clone of the parent in a run of the \ocl. Hence, in those cases the \ocl mimics the \opl and the statement follows since the \ocl finds the target before time $T$ w.h.p.

    For the second case, we observe that a run of the \opl on \OM finds the target in time $O(n \cdot(\lambda + \ln(n/k^\ast)))$ by Theorem~\ref{the:big-ass-fixed-target-theorem}(2), and the second case follows similar to the first case.
\end{proof}
}

\inShortVersion{
For other values of $p$ and $\lambda$, the key step is to show that w.h.p.\ the \opl is at most a factor $O(1/p)$ slower on \disOM than on \OM, which is shown by the following lemma. After a discussion we will show that the upper bounds on $\Tplus$ follows from the domination results in Section~\ref{sec:domination}, using that the \opl on \OM is at least as fast as the \ocl on \disOM. 
 }%
\inLongVersion{The next lemma is the most complicated step. We will show that the local optima in \disOM can increase the runtime of the \ocl at most by a factor of $O(1/p)$. Note that this statement seems very intuitive: the algorithm can always make progress by staying within the distorted points. If the probability of making an improving step on \OM is $p_{\text{imp}}$, then the probability of making an improving step on \disOM should be $p_{\text{imp}}\cdot p$ since we need to find a \OM-improving step, and the offspring needs to be distorted. 

However, this intuition can be misleading. The problem is that the distortions are fixed, and that we do not get fresh randomness each time. Assume we are at distance $k$ from the optimum, and let us focus on single-bit flips for illustration. Since $k$ of the $n$ neighbours are improving, each single-bit flip has a chance of $k/n$ to be improving, so on \onemax we need to wait for $n/k$ single-bit flips in expectation. 

For \disOM, if we are in a distorted point $x$ at distance $k$ from the optimum, then the probability that a single-bit flip is distorted \emph{and} $\OM$-improving (i.e., closer to $\vec 1$ than the parent) is naively $pk/n$, so it is tempting to assume that one simply needs to wait for $n/(pk)$ rounds in expectation. However, this is not true! Once we have queried the fitness of a $\OM$-improving neighbour and it was clean, the chance is gone for good. It could happen that all $k$ $\OM$-improving neighbours of $x$ are clean. In fact, this is \emph{likely} since the expected number of \OM-improving distorted neighbours is $pk$, and $pk =o(1)$ is a perfectly normal situation.\footnote{Recall that $p$ can be arbitrarily close to $1/n$. In fact, this is a particularly interesting case since it gives the largest factor between $\Tcom$ and $\Tplus$ in Theorem~\ref{thm:main}.}
In this case, we can never escape from $x$ by a single-bit flip. (In other words: \disOM has a local optimum, which is the whole point of this new benchmark after all.) So in this case, we need at least \emph{two-bit flips} to escape. 

This could potentially be very costly, but fortunately we can profit from two-bit flips to  \emph{the same $\OM$-level}, i.e., to search points in the same Hamming distance from the optimum. Those two-bit flips are much cheaper than \emph{$\OM$-improving} two-bit flips and provide fresh randomness. Mind that this is a real and important issue, and the following lemma would simply be wrong if the \opl would break ties in favour of the parent. This is also why we are uncertain whether Theorem~\ref{thm:largest-factor} transfers similarly to other functions. In fact, we conjecture that it is false for other linear functions.  
}



\begin{lemma}\label{lem:couple-OM-disOM}
    Consider the setting of Theorem~\ref{thm:largest-factor}, and assume additionally that $p\cdot n\cdot(\lambda + \ln(n/k^\ast))=\Omega(1)$. Let $T^{\sss{\OM}} = T^{\sss{\OM,\plus}}$ be the fixed-target hitting time of  the \opl on \onemax for target fitness $n-k^\ast$, and similarly for $T^{\sss{\disOM}}$. If $T^{\sss{\OM}} \le n^{C'}$ holds w.h.p.\ for some constant $C'>0$, then w.h.p.\
    \begin{align*}
    T^{\sss{\disOM}} = O\big(T^{\sss{\OM}}/p).
    \end{align*}
\end{lemma}
\inShortVersion{This statement seems very intuitive: the algorithm can always make progress by staying within the distorted points. If the probability of making an improving step on \OM is $p_{\text{imp}}$, then the probability of making an improving step on \disOM should be $p_{\text{imp}}\cdot p$ since we need to find a \OM-improving step to a distorted offspring. 

However, this intuition can be misleading. The problem is that the distortions are fixed, and that we do not get fresh randomness each time. Let us focus on single-bit flips for illustration. In distance $k$ from the optimum, since $k$ of the $n$ neighbours are improving, each single-bit flip has a chance of $k/n$ to be improving, so on \onemax we need to wait for $n/k$ single-bit flips in expectation. 

For \disOM, if we are in a distorted point $x$ at distance $k$ from the optimum, then the probability that a single-bit flip is distorted \emph{and} $\OM$-improving (i.e., closer to $\vec 1$ than the parent) is naively $pk/n$, so it is tempting to assume that one simply needs to wait for $n/(pk)$ rounds in expectation. However, this is not true! Once we have queried the fitness of a $\OM$-improving neighbour and it was clean, the chance is gone for good. It could happen that all $k$ $\OM$-improving neighbours of $x$ are clean. In fact, this is \emph{likely} since the expected number of \OM-improving distorted neighbours is $pk$, and $pk =o(1)$ is a perfectly normal situation.\footnote{Recall that $p$ can be arbitrarily close to $1/n$. In fact, this is a particularly interesting case since it gives the largest factor between $\Tcom$ and $\Tplus$ in Theorem~\ref{thm:main}.}
In this case, we can never escape from $x$ by a single-bit flip. (In other words: \disOM has a local optimum, which is the whole point of this new benchmark after all.) So in this case, we need at least \emph{two-bit flips} to escape. 

This could potentially be very costly, but fortunately we can profit from two-bit flips to  \emph{the same $\OM$-level}, i.e., to search points in the same Hamming distance from the optimum. Those two-bit flips are much cheaper than \emph{$\OM$-improving} two-bit flips and provide fresh randomness. Mind that this is a real and important issue, and the above lemma would simply be wrong if the \opl would break ties in favour of the parent. This is also why we are uncertain whether Theorem~\ref{thm:largest-factor} transfers similarly to other functions. In fact, we conjecture that it is false for other linear functions. 

In the proof, we call a distorted search point \emph{good} if at most half of its $\OM$-improving neighbours have been queried before. In this case, the naive intuition is still correct, since each $\OM$-improving neighbour has probability at least $1/2$ to be unqueried, in which case it is distorted with probability $p$. Thus progress within distorted points is only slowed down by a factor of $O(1/p)$. To deal with bad points, we define $N^{(i)}(x)$ as the set of all search points in distance $2i$ from $x$ which can be reached via $i$ consecutive $2$-bit flips, all of which stay within distorted points. Informally, we then show for some constant $i_0$ that w.h.p.\ \emph{(i)} at least half of the search points in $N^{(i_0)}(x)$ are good, and \emph{(ii)} the algorithm moves from $x$ to an almost uniform random point in $N^{(i_0)}(x)$ in expected time $O(1/(p\pimp)+\lambda)$, conditionally on not leaving $\CD_k$. 
This allows us to prove that the algorithm visits good search points frequently, on which it then has good improvement chances. 
We omit the details.
}%
\inLongVersion{
\begin{proof}
W.h.p.\ the intial point of the \opl has distance at least $n/3$ from $\vec 1$. Hence, by Theorem~\ref{the:big-ass-fixed-target-theorem}, w.h.p.\  $T = \Omega(n\ln(n/k^\ast))$.

We first make some observations that allow us to simplify the problem. On \disOM, every search point in Hamming distance $\le n-k^\ast$ from $\vec 1$ has fitness at least $n-k^\ast$. (The converse is not true in general.) Therefore, it suffices to bound the time until the \opl reaches Hamming distance $k^\ast$ from $\vec 1$, since this time is at least as large as $\Tdis$. 

Next, consider a run of the \opl on \disOM. We split $\Tdis = \TdisC + \TdisD$, where $\TdisC$ and $\TdisD$ are the number of function evaluations that are performed while the parent is in a clean and in a distorted state respectively. We will first show that $\TdisC = O(\lambda n + T)$ w.h.p. To see this, we introduce some terminology. We call \emph{level $k$} the set of all search points at distance $k$ from $\vec 1$. We denote the set of clean points on level $k$ by $\CC_k$, and the set of distorted points on level $k$ by $\CD_k$. We call an offspring \emph{$\OM$-improving} if the offspring has strictly smaller distance from $\vec 1$ than the parent. For a parent on fitness level $k$, let $\pimp = \pimp(k)$ be the probability that a mutation creates a $\OM$-improving offspring, and $\pimpone = \pimpone(k)$ be the probability that a mutation is a single-bit flip that creates a $\OM$-improving offspring. Then both $\pimp = \Theta(k/n)$ and $\pimpone = \Theta(k/n)$. 

Let $T_k$ be the number of offspring that the \opl on \disOM generates with parents in $\CC_k$. Note that the algorithm may leave and re-enter $\CC_k$ if there are distorted points of the same fitness. But as soon as it generates a \OM-improving offspring $y$ from a parent $x\in \CC_k$ on level $k$, $y$ is strictly fitter than $x$ (regardless of whether $y$ is distorted or not) and the algorithm leaves $\CC_k$ for good. Hence, every offspring of a parent in $\CC_k$ has probability at least $\pimp$ to leave $\CC_k$ for good. Note that the other offspring in the same generation are still generated, which adds at most $\lambda-1$ additional offspring to $T_k$. Therefore, $T_k$ is stochastically dominated by $T'_k + \lambda-1$, where $T'_k$ follows a geometric distribution $\textrm{Geo}(\pimp(k))$ with $\pimp(k) = \Theta(k/n)$. Summing over all $k$, we may dominate $\TdisC$ by $(\lambda-1)(n-k^\ast)+\sum_{k= k^\ast+1}^{n} T_k' \le \lambda n + \sum_{k= k^\ast+1}^{n} T_k'$ for independent geometric random variables $T_k'$. If $k^\ast \ge n/2$ then w.h.p.\ the sum is $O(n)$ by the Chernoff bound. If $k^\ast < n/2$ then the sum has expectation $\Theta(\sum_{k=k^\ast+1}^n n/k) = \Theta(n\ln(n/k^\ast))$ and is concentrated around its expectation by~\cite[Theorem~1]{witt2014fitness}. In either case, w.h.p.\ $\TdisC = O(\lambda n + n\ln(n/k^\ast))$. \smallskip

Hence, it remains to bound $\TdisD$. 
Let $\TdisDk$ be the time the algorithm spends in the set $\CD_k$. We pessimistically assume that the algorithm enters the set $\CD_k$ for all $k\ge k^\ast$.
Note that, if $d$ is an integer, the algorithm might leave and return to $\CD_k$ by visiting clean points of the same fitness (at level $k+d$). 
We will ignore this complication for now, and only return to it in the very end. 
We will also pessimistically ignore the option that the algorithm might find a clean point of strictly higher fitness, and also ignore the option that the algorithm finds a strictly fitter offspring by flipping several bits at once. Instead, we will assume that none of these options happen (which would only help us), and show that then the algorithm creates a $\OM$-improving distorted offspring by a single-bit flip in expected time $O(T_k/p)$. To do this, the algorithm must \emph{(i)} create a $\OM$-improving neighbour $y$, and \emph{(ii)} $y$ must be distorted. 

Every $\OM$-improving offspring that has not been queried before has probability $p$ to be distorted. If the algorithm queries $2\ln (n)/p$ different $\OM$-improving offspring, then the probability than none of them is distorted is $(1-p)^{2\ln (n)/p} \le e^{-2\ln n} = n^{-2}$. By a union bound over the $\le n$ levels, the probability that this happens for any level is $o(1)$. Hence, we may assume that for all levels, we need to explore at most $2\ln (n) /p$ different $\OM$-improving points from a parent at distance $k$ from $\vec 1$ until we find a distorted one. 

We partition $\CD_k$ into two sets of \emph{good} and \emph{bad} points. The set $\CD_k^{\sss{\text{good}}}$ contains all $x\in \CD_k$ for which the algorithm has so far queried at most $k/2$ of the $k$ $\OM$-improving neighbours of $x$, and $\CD_k^{\sss{\text{bad}}} := \CD_k\setminus \CD_k^{\sss{\text{good}}}$. 

Assume that the algorithm is in $\CD_k^{\sss{\text{good}}}$  and creates a $\OM$-improving neighbour $y$ as offspring. Since all neighbours are equally likely, $y$ has not been queried before with probability at least $1/2$, and in this case it has probability $p$ to be distorted. Hence, in $\CD_k^{\sss{\text{good}}}$, with each offspring the algorithm has probability at least $p_{\text{imp},1}\cdot p/2 = \Theta(p_{\text{imp}}p)$ to find a $\OM$-improving distorted neighbour.

Next assume that the algorithm is in a search point $x\in \CD_k^{\sss{\text{bad}}}$. Recall that we may assume that it has explored at most $2\ln (n)/p$ \OM-improving search points. For $x' \in \CD_k$, let $N_x(x') := \{x''\in \CD_k \mid H(x'',x')=2, H(x'',x) = H(x',x) + 2\}$ be the set of points in $\CD_k$ in distance $2$ of $x'$ that are further away from $x$. Then we define recursively $N^{(1)}(x) := N_x(x) = \{x'\in \CD_k \mid H(x',x)=2\}$ and $N^{(i)}(x) = \bigcup_{x' \in N^{(i-1)}} N_x(x')$ for $i\ge 2$. As we will show, the $N^{(i)}(x)$ constitute a network on which the algorithm may move relatively quickly. Recall the parameter $\eps$ from the condition $k^\ast\ge n^\varepsilon$ in Theorem~\ref{thm:largest-factor}. In the following, we will show that for $i_0 := \lceil 1+1/\eps\rceil$, w.h.p.\
\begin{enumerate}[(i)]
    \item all $N_x(x')$ have size $\Theta(pkn)$, for all $x$ that the algorithm visits and all $x'\in N^{(i_0)}(x)$, 
    \item at least half of the search points in $N^{(i_0)}(x)$ are good, and
    \item informally stated (precise statement below), the algorithm moves from $x$ to a nearly uniform random point in $N^{(i_0)}(x)$ in expected time $O(1/(pp_{\text{imp}})+\lambda)$, conditionally on not leaving $\CD_k$. 
\end{enumerate}
Thus, informally speaking, the algorithm cannot stay in bad search points for long.

We continue by proving the statements in (i), (ii) and (iii), starting with (i). Assume that $i\!\le\! i_0\! =\!O(1)$ and fix $x'\! \in\! N^{(i)}(x)$. Every $x''\!\in\! N_x(x')$ is obtained from $x'$ by flipping a shared zero-bit of $x$ and $x'$ (which gives $k-i$ options) and a shared one-bit ($n-k-i$ options). Moreover, every search point obtained in this way is in $N_x(x')$ if and only if it is distorted, which happens with probability $p$. Hence, $\E[|N_x(x')|] = p\cdot (k-i)(n-k-i) \ge pkn/2$. Since $pkn \ge k/\ln^2 n \ge n^{\eps}/\ln^2 n$, and since $|N_x(x')|$ is binomially distributed, we have $\Prob\big(|N_x(x')| \le pkn/4\big) \le 2^{-pkn/2} \le 2^{-n^{\eps/2}}$ by the Chernoff bound~\cite{DoerrProbabilityChapter2020}. We will now argue that we can afford a union bound over all $x$ and $x'$. Since we assumed $T\le n^{C'}$, we need to use a union bound over at most $n^{C'}$ search points $x$. For each $x$, there are $O(n^i)$ search points at distance $i$ from $x$, which implies $|N^{(i)}(x)| = O(n^i)$. Therefore, we can also afford a union bound over all $x' \in N^{(i)}(x)$ since $i_0=O(1)$, and obtain that w.h.p.\ $|N_x(x')| \ge pkn/4$ holds for all search points $x$ that the algorithm visits, all $x'\in N^{(i)}(x)$ and all $i\le i_0$. By an analogous argument, we also have w.h.p.\ $|N_x(x')| \le 2pkn$ for all such $x$ and $x'$. Note that by the iterative definition of $N^{(i)}(x)$, this implies  $4^{-i}(pkn)^i \le |N^{(i)}(x)| \le 2^i(pkn)^i$. In particular, note that $|N^{(i_0)}(x)| = \Theta((pkn)^{i_0}) = \Omega((n^{\eps}/\ln^2 n)^{i_0}) = \Omega(n)$.

For (ii), let us assume for the sake of contradiction that at least half of $N^{(i_0)}(x)$ is bad. Then each of the bad points $x' \in N^{(i_0)}(x)$ has at least $k/2$ $\OM$-improving neighbours that are already queried. Moreover, at least $k/2 -i_0$ of these neighbours $y$ have distance $2i_0+1$ from $x$. On the other hand, each such $y$ (in distance $k-1$ from the optimum and in distance $2i_0+1$ from $x$) differs from $x$ in exactly $i_0+1$ one-bits and $i_0$ zero-bits, and is therefore neighbour of at most $i_0+1$ points $x' \in N^{(i_0)}(x)$. Hence, the algorithm has queried at least $|N^{(i_0)}(x)|/2 \cdot (k/2 -i_0)/(i_0+1) = \Omega(nk) = \Omega(n^{1+\eps})$ different \OM-improving neighbours. This is a contradiction, since the number of queried $\OM$-improving search points is at most $2\ln (n)/p \le 2n\ln^3 n$. This proves (ii).

For (iii), we first need to give the exact statement. We will show the following for a suitable constant $D>0$. Starting in $x$, fix $x'\in N^{(i_0)}(x)$. Conditional on not leaving $\CD_k$, with probability at least $1/(D |N^{(i_0)}(x)|)$ the point $x'$ will be the first point in $N^{(i_0)}(x)$ that the algorithm visits, and it is visited within time $O(1/(p_{\text{imp}}p))$. Since these events are mutually exclusive for different $x'\in N^{(i_0)}(x)$, this implies in particular that the algorithm visits $N^{(i_0)}(x)$ with probability at least $1/D$ in time $O(1/(p_{\text{imp}}p))$. Moreover, together with (ii) it implies that the algorithm visits a \emph{good} point in $N^{(i_0)}(x)$ with probability at least $1/(2D)$ in time $O(1/(p_{\text{imp}}p))$.


To prove (iii), let us fix some $x'\!\in\! N^{(i_0)}(x)$, and let $x\!=\!x^{(0)}$, $x^{(1)},\ldots,x^{(i_0)}\!=\!x'$ be a chain of search points such that $x^{(i)} \in N_x(x^{(i-1)})$. Let $\CE$ be the event that the $x^{(i)}$ are the next $i_0$ steps in which the algorithm moves to a new search point. Then we will show that 
\[
\Prob(\CE\mid\mbox{not leaving $\CD_k$ for $i_0$ moves}) = \Omega(1/|N^{(i_0)}(x)|).
\]
First note that the probability that the algorithm moves within $\CD_k\subseteq\CD$ in one round is $O(\lambda p p_{\text{imp}})$, since for moving it is necessary to flip at least one zero-bit (which has probability $\Theta(p_{\text{imp}})$), and the offspring needs to be distorted (probability $O(p)$). Finally, the factor $\lambda$ comes from a union bound over the $\lambda$ offspring per generation. 
On the other hand, the probability of moving from $x^{(i)}$ to $x^{(i+1)}$ is $\Theta(\lambda/n^2)=\Theta(\lambda p_{\text{imp}}/(kn) ) = \Theta(\lambda p p_{\text{imp}}/(pkn))$. Hence, the conditional probability of moving to $x^{(i)}$, on moving at all, is $\Theta(1/(pkn))$. Iterating this over the $i_0$ steps (where $i_0$ is a constant), we obtain
\begin{align*}
    \Prob(\CE &\mid\mbox{not leaving $\CD_k$ for $i_0$ moves}) \\&= \prod_{i=0}^{i_0-1} \Theta(1/(pkn)) =  \Theta((pkn)^{-i_0}) \stackrel{(i)}{=} \Theta(1/|N^{(i_0)}(x)|).
\end{align*}
Moreover, each jump from  $x^\sss{(i)}\in N^{(i)}(x)$ to \emph{some} $x^\sss{(i+1)}\in N^{(i+1)}(x)$ has probability $\Omega(1)$ to happen within time $O(1/(pp_\text{imp})+\lambda)$, where the $+\lambda$ is necessary because time (i.e., the number of function evaluations) is always a multiple of $\lambda$. So the probability that each of the $i_0$ steps takes time $O(1/(pp_\text{imp})+\lambda)$ is also $\Omega(1)$. This proves (iii).


It remains to put everything together. If the algorithm is in a good search point, then each offspring has probability $\Omega(p_{\text{imp}}p)$ of being a \OM-improving distorted search point. If the algorithm is in a bad search point, then it has probability $\Omega(1)$ to reach a good search point in time $O(1/(p_{\text{imp}}p)+\lambda)$, if it does not leave $\CD_k$. In any case, in time $O(1/(p_{\text{imp}}p)+\lambda)$ it always has a probability of $\Omega(1)$ of finding a $\OM$-improving distorted neighbour. Finally, if the algorithm leaves $\CD_k$, then this is either due to a fitness improvement which leaves the level directly, or the algorithm goes to a clean point of the same fitness, in which case the chance of improving the fitness is larger. Hence, in all cases we have a chance of $\Omega(1)$ of leaving the fitness level within  time $O(1/(p_{\text{imp}}p)+\lambda)$. In particular, $\E[\TdisDk] = O(1/(p_{\text{imp}}p)+\lambda) = O(\E[T_k]/p + \lambda)$, and hence 
\begin{align*}
\E[\Tdis] & = \E[\TdisC + \TdisD] = O\left(T  + \lambda n + \sum_{k=k^\ast+1}^n T_k/p\right) \\
& = O(\lambda n \ln(n/k^\ast) n+ T/p).
\end{align*}
Moreover, since we can stochastically dominate $\TdisDk$ by independent geometrically distributed random variables, we also get concentration.
We obtain that w.h.p.\ $\Tdis$ is of order at most
\[ 
n\cdot(\lambda + \ln(n/k^\ast))+T/p = O(T/p)
\]
since $p=\Omega(1/(n\cdot(\lambda + \ln(n/k^\ast))))$, and $T\ge 1$.
This concludes the proof. 
\end{proof}
}
Now we have all ingredients to prove Theorem~\ref{thm:largest-factor}. 

\begin{proof}[Proof of Theorem~\ref{thm:largest-factor}]

    Since we switch between the fitness functions $f\in \{\OM,\disOM\}$ and several target fitnesses $k^\ast$, we include the indices $f$ and $k^\ast$ in the notation.


    The first two cases are implied by Lemma~\ref{lem:small-p}. For the third case, we observe that a necessary condition for reaching fitness level $n-k^\ast$ is to reach Hamming distance $k^\ast+d$ from $\vec 1$, so 
    $T^{\sss{\disOM, \mathrm{com}}}_{k^\ast+d}\!\le\! T$.
By Theorem~\ref{thm:couple-com-plus}(a) the \ocl on \disOM is at most as fast as the \opl on \OM, so we also have $T^{\sss{\OM, \plus}}_{k^\ast+d}\!\le\! T$ w.h.p.
    By Theorem~\ref{the:big-ass-fixed-target-theorem} we have $T^{\sss{\OM, \plus}}_{k^\ast} \!=\! \Theta(T^{\sss{\OM, \plus}}_{k^\ast+d})$ w.h.p., where we use $\lambda=O(\ln (n/k^\ast))$ and $k^\ast\le k^\ast+d\le 2k^\ast\le n/3$. 
    Hence, w.h.p.\ $T^{\sss{\OM, \plus}}_{k^\ast}\! =\! O(T)$.
    Then we use Lemma~\ref{lem:couple-OM-disOM} to conclude $T^{\sss{\disOM, \plus}}_{k^\ast}\! =\! O(T/p)$.
    
    We show now that $\Tplus=O(T\cdot n\ln n)$. For parameters as in the first case this is trivial. Assume otherwise, so that $\lambda=O(\ln n)$. If $p\cdot n(\lambda +\ln(n/k^\ast))=o(1)$, then the second case yields $\Tplus=O(n\ln n)=O(T\cdot n\ln n)$. If $p\cdot n(\lambda +\ln(n/k^\ast))=\Omega(1)$, then the third case yields $T=O(T\cdot n(\lambda + \ln(n/k^\ast)))=O(T\cdot n\ln n)$.
\end{proof}

\section{Combining all results}
We verify that the previous sections combined prove Theorem~\ref{thm:main}.
\begin{proof}[Proof of Theorem~\ref{thm:main}]
The upper bound on $\Tcom$ is proven in Section~\ref{sec:com-upper}. 
We verify now the lower bound on $\Tcom$.
    Observe that $\Tcom$ stochastically dominates the time $T'$ until the \ocl finds an offspring $y$ with $\OM(y)\ge k^\ast + d$. By 
Theorem~\ref{the:big-ass-fixed-target-theorem}(1) it follows that $T'=\Omega(n \ln n)$ (setting $\CA=$\ocl, $a=\Theta(n)$ and $b=k^\ast+d$).  

We turn to $\Tplus$. The upper bound follows immediately from the upper bound on $\Tcom$, since by Theorem~\ref{thm:largest-factor} (which holds for a wider range of parameters than assumed in Assumption~\ref{ass:main} in Theorem~\ref{thm:main}) the runtime of the \opl on \disOM is at most a factor $O(1/p)$ slower than on $\OM$ when $\lambda=\Theta(\ln n)$ and $k^\ast=n^{1-\Omega(1)}$. 
The lower bound on $\Tplus$ is given in Section~\ref{sec:plus-lower}.
\end{proof}
\inLongVersion{
Lastly, we give the proof of Proposition~\ref{prop:Markov-Vegas}.
\begin{proof}[Proof of Proposition~\ref{prop:Markov-Vegas}]
For the w.h.p.\ statement in Proposition~\ref{prop:Markov-Vegas}, we just observe that for \onemax, w.h.p.\ both algorithms find the optimum with $O(n\ln n)$ fitness evaluations~\cite{rowe2014choice}. Since $p$ is so tiny, w.h.p.\ none of the $O(n\ln n)$ visited search points is distorted, and thus w.h.p.\ the runtime on \OM and on \disOM is the same. 

Let us now consider $\E[\Tplus]$. With probability $p=2^{-n}$, the all-zero string $\vec 0$ is distorted. With probability $(1-p)^{2^{n}-1} \approx 1/e$, the other $2^{n}-1$ search points are not distorted. So with probability $\Theta(p)$, we have $\vec 0$ as the unique distorted search point. If this happens, then the \opl starts in $\vec 0$ with probability $1/2^n$. If this happens, $\vec 0$ has a fitness of $n-0.5$ and the \opl can only escape by sampling the global optimum. To do that, the algorithm needs to flip all $n$ bits at the same time. The probability to do this is $n^{-n}$, so the algorithm needs expected time $n^{n}$ to escape. In total, this scenario contributes $\Theta(p\cdot 2^{-n} \cdot n^{n}) = n^{\Omega(n)}$ to $\E[\Tplus]$. This proves the lower bound on $\E[\Tplus]$. 

For the \ocl, first note that if the algorithm visits any distorted search point except $\vec 0$, then this search point has fitness at least $n$, so that the fitness target is achieved. Hence, the algorithm terminates when it reaches any distorted search point except $\vec 0$. Therefore, any distorted search point except for $\vec 0$ makes the runtime smaller. We may thus pessimistically assume that all search points except for $\vec 0$ are clean.

This leaves us with two cases. If $\vec 0$ is also clean, then $\disOM = \OM$, and the expected runtime of the \ocl with $\lambda = 3\ln n$ is $O(n\ln n)$ in this case~\cite{rowe2014choice}. In the other case, $\vec 0$ is distorted. Note that this case occurs only with probability $p=2^{-n}$. If the algorithm does not hit $\vec 0$, then we can argue as before, so let us pessimistically assume that the algorithm starts in $\vec 0$. Then after an expected $\Theta(q^{-1})=  n^{O(1)}$ rounds, it does not duplicate $\vec 0$ and thus proceeds to a search point $x\neq \vec 0$. Thus, there is an $i$ such that $x_i \neq 0$. We claim that the \ocl has a probability of at least $\rho = n^{-O(1)}$ to reach the optimum in $n^{O(1)}$ steps from $x$ without visiting $\vec 0$ again. Note that this implies the bound on $\E[\Tcom]$, since whenever the \ocl is in $\vec 0$, it has a probability of $\Omega(\rho)$ of leaving $\vec 0$ and reaching the optimum in the next $n^{O(1)}$ steps. Hence, this case contributes at most $p \cdot O(1/\rho)\cdot n^{O(1)}=2^{-n}n^{O(1)} \le 1$ to $\E[\Tcom]$.

So it remains to prove the estimate for $\rho$. Let $C>0$ be a large constant to be fixed later. With probability $(1-1/n)^{C n\ln n} \ge e^{-2C\ln n} = n^{-2C}$, the \ocl does not flip the $i$th bit in any of the next $2C n \ln n$ mutations. If this happens, then the algorithm in particular does not return to $\vec 0$ during this time. Moreover, if the $i$th bit remains unchanged then the algorithm simply optimizes an $(n-1)$-dimensional \onemax instance during this time.\footnote{Note that for a \onemax problem on $n' \coloneqq n-1$ bits, the mutation rate is $1/n=1/(n'+1)$ instead of $1/n'$. However, it is clear that this deviation is negligible.} If $C$ is sufficiently large, the  expected time to reach the optimum of the $(n-1)$-dimensional problem is at most $Cn\ln n$, and by Markov's inequality the probability of needing more than $2Cn\ln n$ steps is at most $1/2$. Hence, the probability of finding the optimum in $Cn\ln n$ rounds is at least $\tfrac12 n^{-2C}$.
This concludes the proof. 
\end{proof}
}

\section{Conclusion}
We have shown that a comma strategy can indeed help for dealing with local optima. To this end, we have introduced the new theoretical benchmark \disOM. 
We believe that this benchmark is of wider interest for studying local optima. As discussed in the introduction, arguably the popular benchmarks \jump and \cliff have rather atypical local optima, and \disOM is a very simple way of adding local optima to the simple \OM function. Thus, it would be very interesting to investigate how other non-elitist selection mechanisms like tournament selection~\cite{lehre2022more}, linear ranking selection, or fitness-proportionate selection~\cite{happ2008rigorous} perform (see~\cite{goldberg1991comparative,lehre2011fitness} for overviews on non-elitist selection), and whether this can be phrased more generally in terms of selective pressure~\cite{lehre2010negative}. \inLongVersion{It would also be interesting to see whether this allows for parameter settings that find the optimum on \disOM efficiently, rather than only reaching a fixed target.}

Our proof also gives insights into how the \ocl escapes local optima. In particular, in the \disOM landscape under Assumption~\ref{ass:main}, our proof shows that the \ocl escapes local optima for good: after escaping, it never hits the same local optimum a second time. 

Of course, \onemax is not the only function that can be distorted. The same process can be applied to any other function, for example to any linear function. As discussed 
\inShortVersion{after} %
\inLongVersion{before}%
Lemma~\ref{lem:couple-OM-disOM}, we suspect that for the \opl there is a real difference between \onemax and other linear functions, and that the huge fitness plateaus of \onemax are important for the \opl to be efficient.

\begin{acks}
We are thankful for the fruitful discussions at the Dagstuhl seminar 22081 ``Theory of Randomized Optimization Heuristics'', which triggered this research, as well as the Dagstuhl seminar 22182 ``Estimation-of-Distribution Algorithms: Theory and Applications''.
\end{acks}

\balance
\bibliographystyle{ACM-Reference-Format}
\bibliography{references}



\end{document}